\documentclass{article}

\usepackage{microtype}
\usepackage{graphicx}
\usepackage{subfigure}
\usepackage{booktabs} 

\usepackage{hyperref}
\usepackage{xcolor,soul}
\usepackage{url}
\usepackage{wrapfig}
\usepackage{lipsum}  
\usepackage{listings} 

\usepackage{amssymb}
\usepackage{pifont}
\usepackage{multirow}
\usepackage{multicol}
\usepackage{makecell}
\usepackage{arydshln}

\usepackage[accepted]{icml2024}

\usepackage{amsmath}
\usepackage{amssymb}
\usepackage{mathtools}
\usepackage{amsthm}

\usepackage[capitalize,noabbrev]{cleveref}
\usepackage{caption}
\usepackage{subcaption,siunitx,booktabs}

\theoremstyle{plain}

\theoremstyle{definition}

\theoremstyle{remark}

\usepackage[textsize=tiny]{todonotes}

\definecolor{MutedRed}{RGB}{255, 0, 0} 
\definecolor{MutedBlue}{RGB}{0, 0, 255} 
\definecolor{darkgreen}{rgb}{0.0, 0.3, 0.0}
\definecolor{darkred}{rgb}{0.3, 0.0, 0.0}
\newcommand{\bR}{\color{MutedRed}\boldmath}
\newcommand{\bB}{\color{MutedBlue}}

\definecolor{moh_colour}{RGB}{255, 204, 204}

\newcommand{\cmark}{\ding{51}}%
\newcommand{\xmark}{\ding{55}}%
\definecolor{moh_colour}{RGB}{255, 204, 204}

\newcommand{\GN}{FLAN}

\usepackage{array}
\newcolumntype{H}{>{\setbox0=\hbox\bgroup}c<{\egroup}@{}}

\icmltitlerunning{Encodings for Prediction-based Neural Architecture Search}

\begin{document}

\twocolumn[
\icmltitle{Encodings for Prediction-based Neural Architecture Search}



\icmlsetsymbol{equal}{*}

\begin{icmlauthorlist}
\icmlauthor{Yash Akhauri}{yyy}
\icmlauthor{Mohamed S. Abdelfattah}{yyy}
\end{icmlauthorlist}

\icmlaffiliation{yyy}{Cornell University, New York, USA}

\icmlcorrespondingauthor{Yash Akhauri}{ya255@cornell.edu}

\icmlkeywords{Machine Learning, ICML}

\vskip 0.3in
]



\printAffiliationsAndNotice{}  

\begin{abstract}
Predictor-based methods have substantially enhanced Neural Architecture Search (NAS) optimization. The efficacy of these predictors is largely influenced by the method of encoding neural network architectures. 
While traditional encodings used an adjacency matrix describing the graph structure of a neural network, novel encodings embrace a variety of approaches from unsupervised pretraining of latent representations to vectors of zero-cost proxies. 
In this paper, we categorize and investigate neural encodings from three main types: structural, learned, and score-based.
Furthermore, we extend these encodings and introduce \textit{unified encodings}, that extend NAS predictors to multiple search spaces. 
Our analysis draws from experiments conducted on over 1.5 million neural network architectures on NAS spaces such as NASBench-101 (NB101), NB201, NB301, Network Design Spaces (NDS), and TransNASBench-101. 
Building on our study, we present our predictor \textbf{\GN}: \textbf{Fl}ow \textbf{A}ttention for \textbf{N}AS. 
\GN{} integrates critical insights on predictor design, transfer learning, and \textit{unified encodings} to enable more than an order of magnitude cost reduction for training NAS accuracy predictors. Our implementation and encodings for all neural networks are open-sourced at \href{https://github.com/abdelfattah-lab/flan_nas}{https://github.com/abdelfattah-lab/flan\_nas}.
\end{abstract}

\section{Introduction}

In recent years, Neural Architecture Search (NAS) has emerged as an important methodology to automate neural network design.
NAS consists of three components: (1) a neural network search space that contains a large number of candidate Neural Networks (NNs), (2) a search algorithm that navigates that search space, and (3) optimization objectives such as NN accuracy and latency. 
A key challenge with NAS is its computational cost, which can be attributed to the sample efficiency of the NAS search algorithm, and the cost of evaluating each NN candidate.
A vast array of search algorithms have been proposed to improve NAS sample efficiency, ranging from reinforcement learning \citep{rl_nas}, to evolutionary search \citep{enas}, and differentiable methods \citep{liu2018darts}.
To reduce the evaluation cost of each NN candidate, prior work has utilized reduced-training accuracy \citep{econas} zero-cost proxies \citep{abdelfattah2021zero}, and accuracy predictors that are sometimes referred to as surrogate models \citep{nasbench301}. 
One of the most prevalent sample-based NAS algorithms utilizes accuracy predictors to both evaluate a candidate NN, and to navigate the search space.
Recent work has clearly demonstrated the versatility and efficiency of \textit{prediction-based NAS} \citep{brpnas,help}, highlighting its importance.
In this paper, we focus on understanding the makings of an efficient accuracy predictor for NAS, and we propose improvements that significantly enhance its sample efficiency and generality.

An integral element within NAS is the encoding method to represent NN architectures. 
Consequently, an important question arises, \textit{how can we encode NNs to improve NAS efficiency?} 
This question has been studied in the past by \citet{encodingstudy}, investigating the effect of graph-based encodings such as adjacency matrices or path enumeration to represent NN architectures.
However, recent research has introduced a plethora of new methods for encoding NNs which rely on concepts ranging from unsupervised auto-encoders, zero-cost proxies, and clustering NNs by computational similarity to learn latent representations. 
\textit{This motivates an updated study on NN encodings for NAS to compare their relative performance and to elucidate the properties of effecive encodings to improve NAS efficiency.}

\begin{figure}[t!]
    \centering
    \includegraphics[width=\columnwidth]{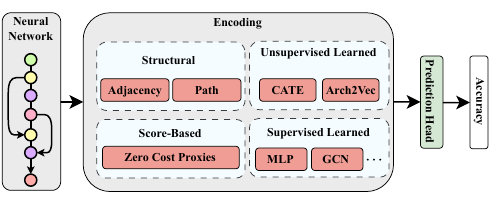}
    \caption{The basic structure of an accuracy predictor highlights that many different types of encodings can be fed to the same prediction head to perform accuracy prediction.}
    \label{fig:main_figures}
\end{figure}
We identify three key categories of encodings. \textbf{Structural} encodings~\citep{encodingstudy} represent the graph structure of the NN architecture in the form of an adjacency matrix or path enumeration, typically represented by an operation matrix to identify the operation at each edge or node. 
\textbf{Score-based} \cite{multipredict} encodings map architectures to a vector of \textit{measurements} such as Zero-Cost Proxies~\citep{snip, synflow, naswot}. 
Finally, \textbf{Learned} encodings learn latent representations of the architecture space. They can be further bifurcated into ones that explicitly learn representations through large-scale unsupervised training \citep{arch2vec, cate} and ones that co-train neural encodings during the supervised training of an accuracy predictor~\citep{tagates, nat_gcn}.
Figure~\ref{fig:main_figures} illustrates our taxonomy of encoding methods, and their role within a NAS accuracy predictor.

\comment{Encodings play an important role in enhancing the performance and efficiency of Neural Architecture Search (NAS), for instance, CATE (\cite{cate}) uses computational similarity between architectures as a way to learn latent representations. Using the CATE encoding could thus convey to the predictor more information about the architecture that the connectivity pattern cannot capture. TAGATES and MultiPredict (\cite{tagates, multipredict}) use zero-cost proxies as a way to encode NNs, which provides a richer representation of the computations in the NN. These encodings lay down a robust foundation for the advent of predictor-based NAS. This branch of NAS, particularly, has been extensively explored through various predictor models, including but not limited to MLPs and GCNs (\cite{multipredict,tagates}). These predictors employ encodings to efficiently model the search space and rapidly identify optimal neural architectures, significantly reducing the computational resources required in comparison to traditional search methods. Notably, these predictors often implicitly learn encodings, acquiring latent representations that grasp important design aspects of the architectural search space. It thus becomes crucial to investigate \textit{what characterizes an effective  predictor}.
}

NN encodings are particularly important in the case of prediction-based NAS because they have a large impact on the effectiveness of training an accuracy predictor.
For that reason, and due to the increasing importance of predictors within NAS, our work provides a comprehensive analysis of the impact of encodings on the sample-efficiency of NAS predictors. 
We validate our observations on 13 NAS design spaces, spanning 1.5 million neural network architectures across different tasks and data-sets.
%
%
Furthermore, NAS predictors have the capability to \textit{extend} beyond a single NAS search space through transfer learning \cite{gennape, cdp_crossdomain} or more generally metalearning \cite{help}. This involves pre-training a predictor on an available NAS benchmark, then efficiently transferring it to a new search space with few NN accuracy samples. Our study examines the role of encodings and transfer learning in predictor efficacy. \textbf{Our Contributions} are:

\begin{enumerate}
\itemsep0em 
    \item We categorize and study the performance of several NN encoding methods in NAS accuracy prediction across 13 different NAS spaces. 
    \item We propose a new hybrid encoder (called \GN{}) that outperforms prior methods consistently on multiple NAS benchmarks. We demonstrate a 2.12$\times$ improvement in NAS sample efficiency.
    \item We create \textit{unified} encodings that allow few-shot transfer of accuracy predictors to new NAS spaces. Notably, we are able to improve sample efficiency of predictor training by 46$\times$ across three NAS spaces compared to trained-from-scratch predictors from prior work.
    \item We generate and provide open access to structural, score-based, and learned encodings for over 1.5 million NN architectures, spanning 13 distinct NAS spaces.
\end{enumerate}

\section{Related Work}

\textbf{Predictor-based NAS.} 
NAS consists of an evaluation strategy to fetch the accuracy of an architecture, and a search strategy to explore and evaluate novel architectures. 
Predictor-based NAS involves training an accuracy predictor which guides the architectural sampling using prediction scores of unseen architectures \citep{brpnas,whiteperfpredictornas}. 
Recent literature has focused on the sample efficiency of these predictors, with BONAS \citep{bonas} using a GCN for accuracy prediction as a surrogate function of Bayesian Optimization, and BRP-NAS \cite{brpnas} employing a binary relation predictor and iterative sampling strategy. 
Recently, TA-GATES \citep{tagates} employed learnable operation embeddings and introduced a method of updating embeddings akin to the training process of a NN to achieve state-of-the-art sample efficiency. 

\textbf{NAS Benchmarks.} 
To facilitate NAS research, a number of NAS benchmarks have been released, both from industry and academia \citep{nasbench101,nasbench301,transnasbench,nbsuite_naseval}. 
These benchmarks contain a NAS space and accuracy for architectures on a specific task.
Even though most of these benchmarks focus on cell-based search spaces for image classification, they greatly vary in size (4k -- 400k architectures) and NN connectivity.
Additionally, more recent benchmarks have branched out to include other tasks~\citep{nbasr} and macro search spaces~\citep{blox}.
Evaluations on a number of these benchmarks have become a standard methodology to test and validate NAS improvements without incurring the large compute cost of performing NAS on a new search space.

\begin{figure*}[t!]
    \centering
    \includegraphics[width=1.8\columnwidth]{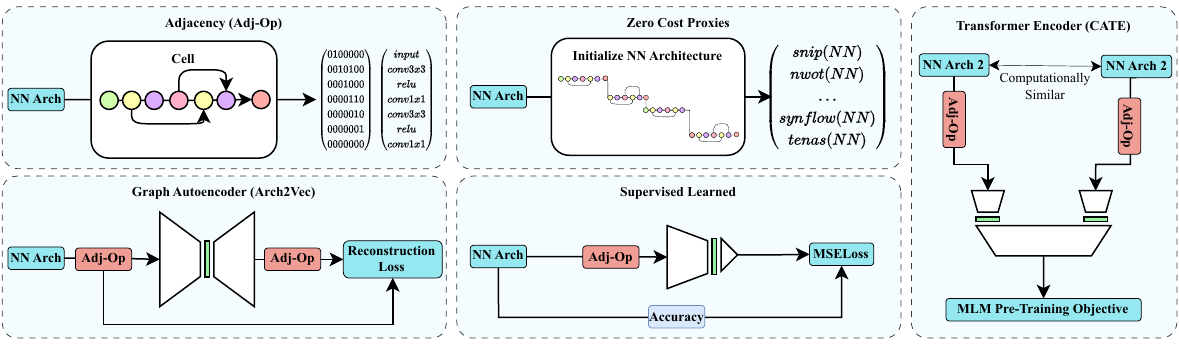}
    \caption{Illustration of important encoding methods that are discussed and evaluated in our work.}
    \label{fig:encoder_types}
\end{figure*}

\textbf{NN Encodings.} 
There are several methods for encoding candidate NN architectures. 
Early NAS research focused on \textit{structural encodings}, converting the adjacency and operation matrix representing the DAG for the candidate cell into a flattened vector to encode architectures \citep{encodingstudy}. 
\textit{Score-based methods} such as Multi-Predict \citep{multipredict} focus more on capturing broad architectural properties, by generating a vector consisting of zero-cost proxies and hardware-latencies to represent NNs. 
There have also been efforts in \textit{unsupervised learned} encodings such as Arch2Vec~\citep{arch2vec}, which leverages the graph auto-encoders to learn a compressed latent vector used for encoding a NN. 
Another method, CATE \citep{cate}, leverages concepts from masked language modeling to learn latent encodings using computational-aware clustering of architectures using Transformers. 
Finally, many supervised accuracy predictors implicitly learn encodings as intermediate activations in the predictor. 
These \emph{supervised learned} encodings have been generated most commonly with graph neural networks (GNNs) within accuracy predictors~\citep{brpnas, bonas, tagates,cdp_crossdomain}.

\section{Encodings}

A basic NAS formulation aims to maximize an objective function $\ell:A\rightarrow\mathbb{R}$, where $\ell$ is a measure of NN accuracy for our purposes but can include performance metrics as well such as hardware latency~\citep{brpnas}. $A$ is a NN search space.
During NAS, NN architectures $a\in A$ are \textit{encoded} using some encoding function $e:A\rightarrow\mathbb{R}^d$, that represents a NN architecture as a d-dimensional tensor.
While prior work \citep{encodingstudy} only considered a narrow definition of encodings wherein $e$ was a fixed transformation that was completely independent of $\ell$, we expand the definition to also consider encoding functions that are parameterized with $\theta$. 
This includes supervised training to minimize the empirical loss $\mathcal{L}$ on predicted values of $\ell$ to actual measurements: $\min_{\theta,\phi}\sum_{a\in A'}\mathcal{L}(F_\phi(e_\theta(a)),\ell(a))$, where $F:E\rightarrow\mathbb{R}$ is a prediction head that takes a learned encoding value $e(a)$ and outputs predicted accuracy $\ell'(a)$.
Simply put, this allows us to evaluate part of an accuracy predictor as a form of encoding, for example, a learned graph neural network encoding function that is commonly used in predictor-based NAS~\citep{tagates}.
Our definition also includes the use of unsupervised training to learn a latent representation $r$, for example using an autoencoder which attempts to optimize $\min_{\theta,\phi}\sum_{a\in A'}\mathcal{L}(F^{enc}_\theta(e(a)),F^{dec}_\phi(r))$ using an encoder-decoder structure that is trained to recreate the graph-based structure of an NN (e.g. adjacency encoding)~\citep{arch2vec}.
Our broader definition of encodings allows us to compare many methods of NN encodings that belong to the four categorizations below---important encodings are illustrated in Figure~\ref{fig:encoder_types}.

\textbf{Structural} encodings capture the connectivity information of a NN exactly. 
\citet{encodingstudy} investigate two primary paradigms for structural encodings, \emph{Adjacency} and \emph{Path} encodings. A neural network can have $n$ nodes, the adjacency matrix simply instantiates a $n \times n$ matrix, where each nodes connectivity with the other nodes are indicated. 
On the other hand, Path encodings represent an architecture based on the set of paths from input-to-output that are present within the architecture DAG. 
There are several forms of these encodings discussed further by \citet{encodingstudy}, including Path truncation to make it a fixed-length encoding. 
Their investigation reveals that Adjacency matrices are almost always superior at representing NNs.

\textbf{Score-based} encodings represent a neural network as a vector of measurements related to NN activations, gradients, or properties. 
These metrics were defined to be a vector of Zero-Cost Proxies (ZCPs) and hardware latencies (HWL) in MultiPredict \citep{multipredict}, and used for accuracy and latency predictors respectively. 
Zero-cost proxies aim to find features of a NN that correlate highly with accuracy, whereas hardware latencies are fetched by benchmarking the architecture on a set of hardware platforms. 
Naturally, connectivity and choice of operations would have an impact on the final accuracy and latency of a model, therefore, these encodings implicitly capture architectural properties of a NN architecture, but contain no explicit structural information. 

\begin{figure*}[t!]
    \centering
    \includegraphics[width=1.7\columnwidth]{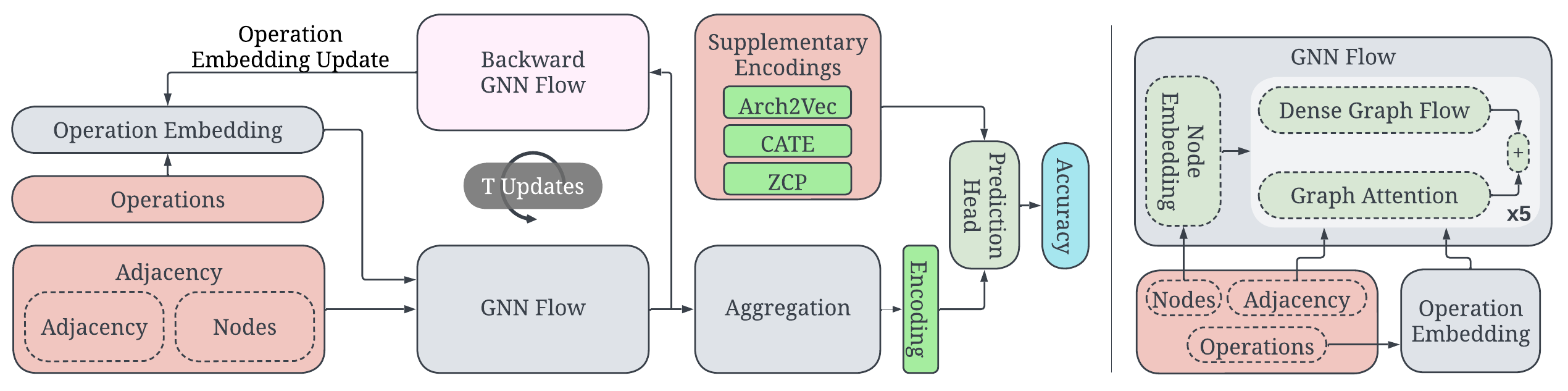}
    \caption{The FLAN predictor architecture showing dual graph flow mechanisms, independent updates of operation embeddings, and the capability to concatenate supplementary encodings.}
    \label{fig:our_network_arch}
\end{figure*}

\textbf{Unsupervised Learned} encodings are representations that aim to distil the structural properties of a neural architecture to a latent space without utilizing accuracy. 
Arch2Vec \citep{arch2vec} introduces a variational graph isomorphism autoencoder to learn to regenerate the adjacency and operation matrix. 
CATE \citep{cate} is a transformer based architecture that uses computationally similar architecture pairs (FLOPs or parameter count) to learn encodings. 
With two computationally similar architectures, a transformer is tasked to predict masked operations for the pairs, which skews these encodings to be similar for NNs with similar computational complexity.
Unsupervised Learned encodings are typically trained on a large number of NNs because NN accuracy is not used.

\textbf{Supervised Learned} encodings refer to representations that are implicitly learned in a supervised fashion as a predictor is trained to estimate accuracy of NN architectures. These encodings are representations that evolve and continually adapt as more architecture-accuracy pairs are used to train an accuracy predictor. 
Supervised Learned encodings are more likely to exhibit a high degree of bias towards the specific task on which they are trained, potentially limiting their generality when extending/transferring a predictor to a different search space.


\begin{table}[t!]
\centering
\resizebox{\columnwidth}{!}{%
      \begin{tabular}{llccccc}
          \toprule
          \textbf{Forward} & \textbf{Backward}                        & \textbf{NB101} & \textbf{NB201} & \textbf{NB301}                  & \textbf{PNAS}                & \textbf{Amoeba}            
          \\\cmidrule(lr){1-2}\cmidrule(lr){3-3}\cmidrule(lr){4-4}\cmidrule(lr){5-5}\cmidrule(lr){6-6}\cmidrule(lr){7-7} \addlinespace[0.5ex]
          \textbf{DGF}     & \textbf{DGF}                               & $0.71$      & $0.80$      &   $0.71$  &\bB$0.38$ & \bB$0.42$ \\
          \textbf{GAT}     & \textbf{GAT}                               & $0.65$      & $0.77$      &   $0.79$  & $0.36$   & $0.38$ \\
          \textbf{Ensemble} & \textbf{DGF}                              &\bB$0.72$    &\bB$0.81$  &\bB$0.81$  &  $0.30$  & $0.39$ \\
          \textbf{Ensemble} & \textbf{Ensemble}       &\bR$0.73$    & \bR$0.82$   &\bR$0.82$  &\bR$0.42$ & \bR$0.46$ \\
          \bottomrule
      \end{tabular}%
  }
\caption{Ensembling both DGF and GAT modules improves predictor performance. Table shows Kendall-$\tau$ coeff. of accuracy predictors trained on 128 NNs and tested on the remainder of each search space.}
\label{tab:module_design} 
\end{table}

\textbf{Unified Encodings.} Multi-Predict \citep{multipredict} and GENNAPE \citep{gennape} introduce encodings that can represent arbitrary NNs across multiple search spaces. Further, CDP \cite{cdp_crossdomain} introduces a predictor trained on existing NAS benchmark data-sets, and is then used to find architectures in large-scale search spaces.
In our work, we look at unified encoding methods that can work across cell-based search spaces to enable NAS knowledge reuse, and to enable search on novel search spaces with only a few samples. 
To make our encodings unified, we append unique numerical indices to the cell-based encoding of each search space. 
This simple extension enables the use of our studied encodings across multiple search spaces.

\section{\GN: Flow Attention Networks For NAS}

Our empirical evaluation (yet to be presented in Table~\ref{tab:motivate_arch}) shows that Supervised Learned encoders often out-perform other encoding methods. 
This is somewhat expected because they have access to the accuracies of NNs in the search space. 
However, training candidate NN architectures can be fairly expensive, and it is not always feasible to obtain accuracies of a sufficient number of NNs, therefore, we focus on the sample efficiency of the accuracy predictors. 
In this section, we introduce \textbf{\GN}: a hybrid encoding architecture which draws on our empirical analysis to deliver state-of-the-art sample efficiency for accuracy prediction.
We carefully tune the predictor architecture so that it can be used reliably as a vehicle to investigate and compare existing and new hybrid encoding schemes as well as unified encodings.
Figure~\ref{fig:our_network_arch} shows the FLAN architecture, described further in this section.
FLAN combines successful ideas from prior graph-based encoders \citep{brpnas,tagates} and further improves upon them through dual graph-flow mechanisms.
In addition to learning an implicit NN encoding, FLAN can be supplemented with additional encodings arbitrarily through concatenation before the predictor head as shown in Figure~\ref{fig:our_network_arch}.

\subsection{GNN Architecture}
Compared to Multi-Layer Perceptrons (MLPs), Graph Convolutional Networks (GCN) improve prediction performance, as shown in Table \ref{tab:motivate_arch}. We employ an architectural adaptation inspired by \citep{gcn_residual_gcnii}, referred to as `Dense Graph Flow' (DGF) \citep{gates}. Empirical analysis, detailed in Table \ref{tab:nb201_arch_design}, reveals that substantial enhancements in predictor performance can be realized through the integration of residual connections \citep{kipf2017semisupervised_gcnarch} within DGF. Further, we add another node propagation mechanism based on graph attention to facilitate inter-node interaction. Empirical results in Table~\ref{tab:module_design} and Table \ref{tab:module_design2} shows that the ensemble of both graph flows (DGF+GAT) typically yields the best results.

\textbf{Dense Graph Flow (DGF): } 
DGF employs residual connections to counteract over-smoothing in GCNs, thereby preserving more discriminative, localized information. 
Formally, given the input feature matrix  for layer $l$ as \(X^{l} \), the adjacency matrix \(A\), and the operator embedding \(O\), with parameter and bias as \(W^{l}_{o} \), \(W^{l}_{f}\), and \(b^{l}_{f}\) respectively, the input feature matrix for the $(l+1)^{th}$ layer is computed as follows (\( \sigma \) is the sigmoid activation):
\begin{equation}
X^{l+1} = \sigma(OW^{l}_{o}) \odot (A (X^{l}W^{l}_{f})) + (X^{l}W^{l}_{f}) + b^{l}_{f}
\label{eq:dgf}
\end{equation}

\textbf{Graph Attention (GAT): } 
Unlike DGF, which employs a linear transform $W^{l}_{o}$ to apply learned attention to the operation features, GAT~\citep{gat_bengio} evaluates pairwise interactions between nodes through an attention layer during information aggregation. The input to the $l^{th}$ layer is a set of node features (input feature matrix) $X^{l}$, to transform the input to higher level-features, a linear transform paramterized by the projection matrix $W^{l}_{p}$ is applied to the nodes. This is followed by computing the self-attention for the node features with a shared attentional mechanism $a$. LR indicates LeakyReLU. The output $X^{l+1}$ is thus calculated as follows:

\begin{equation}
\text{Attn}_{j}(X^{l}) = \sigma(\text{LR}(A_{j} \cdot a(W^{l}_{p}X^{l} \cdot W_{p}X^{l}_{j}))) \cdot W_{p}X^{l}_{j}
\label{eq:attention-coefficients}
\end{equation}
\begin{equation}
X^{l+1} = \text{{LayerNorm}}\left( \sigma(OW^{l}_{o}) \odot \sum_{j=1}^{n} \text{Attn}_{j}(X^{l}) \right)
\label{eq:graph-attention-output}
\end{equation}

where \( \text{Attn}_{j} \) are the normalized attention coefficients, \( \sigma \) denotes the sigmoid activation function. To optimize the performance of GATs, we incorporate the learned operation attention mechanism $W_{o}$ from Equation \ref{eq:dgf} with the pairwise attention to modulate the aggregated information and LayerNorm to improve stability during training.

\begin{table*}[t!]
  \centering
  \resizebox{\linewidth}{!}{%
    \begin{tabular}{llcccccccccccc}\toprule
      \multirow{4}{*}{\bf Classification} & \multirow{4}{*}{\bf Encoder}   & \multicolumn{3}{c}{\textbf{NASBench-101}} & \multicolumn{3}{c}{\textbf{NASBench-201}} & \multicolumn{3}{c}{\textbf{NASBench-301}} & \multicolumn{3}{c}{\textbf{ENAS}} \\
      &                                                               & \multicolumn{3}{c}{(Portion of 7290 samples)}                  & \multicolumn{3}{c}{(Portion of 7813 samples)}             & \multicolumn{3}{c}{(Portion of 5896 samples)}                 & \multicolumn{3}{c}{(Portion of 500 samples)}               \\\cmidrule(lr){3-5} \cmidrule(lr){6-8} \cmidrule(lr){9-11} \cmidrule(lr){12-14} \addlinespace[0.5ex]
                   &                                                  & 1\%                  & 5\%                  & 10\%                 & 0.1\%                & 0.5\%                & 1\%             & 0.5\%                & 1\%                  & 5\%                          & 5\%                  & 10\%                 & 25\%             \\\cmidrule(lr){1-1}\cmidrule(lr){2-2}\cmidrule(lr){3-5} \cmidrule(lr){6-8} \cmidrule(lr){9-11} \cmidrule(lr){12-14} \addlinespace[0.5ex]
\multirow{2}{*}{\textbf{Structural}}      & \textbf{ADJ}                            & $0.327$             & $0.464$             & $0.514$             & $0.047$             & $0.273$             & $0.382$        & 0.275               & 0.401               & 0.537                       & $0.057$             & $0.060$             & $0.089$         \\
& \textbf{Path}                           & 0.387               & 0.696              & 0.752                & 0.133               & 0.307               & 0.396          &   -                  &         -            &       -                      & -                    & -                    & -                \\\cmidrule(lr){1-1}\cmidrule(lr){2-2}\cmidrule(lr){3-5} \cmidrule(lr){6-8} \cmidrule(lr){9-11} \cmidrule(lr){12-14} \addlinespace[0.5ex]
\textbf{Score}                            & \textbf{ZCP}                            &   0.591             &   0.662             &   0.684             &   0.248             &   0.397             &   0.376        &    $0.286$          &$0.272$              &   $0.367$                   &    $0.387$          &   0.458             &   0.540         \\\cmidrule(lr){1-1}\cmidrule(lr){2-2}\cmidrule(lr){3-5} \cmidrule(lr){6-8} \cmidrule(lr){9-11} \cmidrule(lr){12-14} \addlinespace[0.5ex]
\textbf{Unsupervised}                     & \textbf{Arch2Vec}                       & $0.210$             & $0.346$             & $0.345$             & $0.046$             & $0.165$             & $0.144$        &   $0.174$           &    $0.228$          &    $0.379$                  & $0.202$             & $0.228$             & $0.324$         \\
\textbf{Learned}                          & \textbf{CATE}                           & $0.362$             & $0.458$             & $0.467$             & $0.462$             & $0.551$             & $0.571$        &    $0.388$          &    $0.349$          &   $0.417$                   & $0.200$             & $0.279$             &  $0.410$        \\\cmidrule(lr){1-1}\cmidrule(lr){2-2}\cmidrule(lr){3-5} \cmidrule(lr){6-8} \cmidrule(lr){9-11} \cmidrule(lr){12-14} \addlinespace[0.5ex]
\textbf{Supervised} & \textbf{GCN}      &  0.366              &  0.597              &  0.692              &  0.246              &  0.311              &  0.408         &   0.095             &  0.128              &  0.267                      &  0.230              &  0.314              &  0.428          \\
\textbf{Learned} & \textbf{GATES}       &  0.632              &  0.749              &  0.769              &  0.430              &  0.670              &  0.757          & 0.561               & 0.606               & 0.691                 &  0.340              &  0.428              &  0.527          \\
&         \textbf{\GN}                  & $0.665$             &  $0.794$            &  $0.823$            &  $0.486$            & $0.706$             &  $0.782$        &  0.539              &     0.537           &   0.698               &  $0.146$            &  $0.291$            &  $0.505$        \\\cmidrule(lr){1-1}\cmidrule(lr){2-2}\cmidrule(lr){3-5} \cmidrule(lr){6-8} \cmidrule(lr){9-11} \cmidrule(lr){12-14} \addlinespace[0.5ex]
\multirow{5}{*}{\textbf{Hybrid}}          
&           \textbf{TAGATES}            &  0.668              &  0.774              &  0.783              &\bR$0.538$           &  0.670              &  0.773          & \bB$0.572$               & \bB$0.635$               &  \bB$0.712$                &   $0.345$           &  \bB$0.440$              &  0.548          \\
&        \textbf{\GN}$_{ZCP}$           &\bR$0.698$           &\bR$0.811$           &\bR$0.831$           &  $0.510$            &\bR$0.714$           &\bR$0.788$       & \bR$0.573$          &  \bR$0.656$         & \bR$0.721$            &\bR$0.397$           &\bR$0.470$           &\bR$0.589$       \\ 
&        \textbf{\GN}$_{Arch2Vec}$      & $0.609$             &  $0.775$            &  $0.816$            &  \bB$0.524$            & \bB$0.713$             &   \bB$0.785$       &  $0.417$            &  $0.509$            &      $0.688$          &  $0.128$            &  $0.243$            &  $0.410$        \\ 
&        \textbf{\GN}$_{CATE}$          & $0.668$             &  $0.795$            &  $0.827$            &  $0.496$            & $0.694$             &  $0.778$        & $0.527$             &  $0.502$            &       $0.702$         &  $0.172$            &  $0.308$            &  $0.466$        \\  
&        \textbf{\GN}$_{CAZ}$           & \bB$0.689$             &  \bB$0.807$            &  \bB$0.831$            &  $0.489$            & $0.703$             &  $0.782$        &   $0.517$           &  $0.537$            &        $0.698$        &  \bB$0.355$            &  $0.433$            &  \bB$0.570$        \\
\bottomrule
\end{tabular}%
}
  \caption{A comparative study of accuracy predictors when utilizing different encoding methods. Table shows Kendall-$\tau$ correlation coefficient of predictors relative to ground-truth NN accuracies. FLAN$_X$ refers to the FLAN encoder with supplemental X encodings.}
  \label{tab:motivate_arch} 
  \end{table*}

\begin{figure*}[t!]
    \centering
    \vspace{-3mm}
    \includegraphics[width=2 \columnwidth]{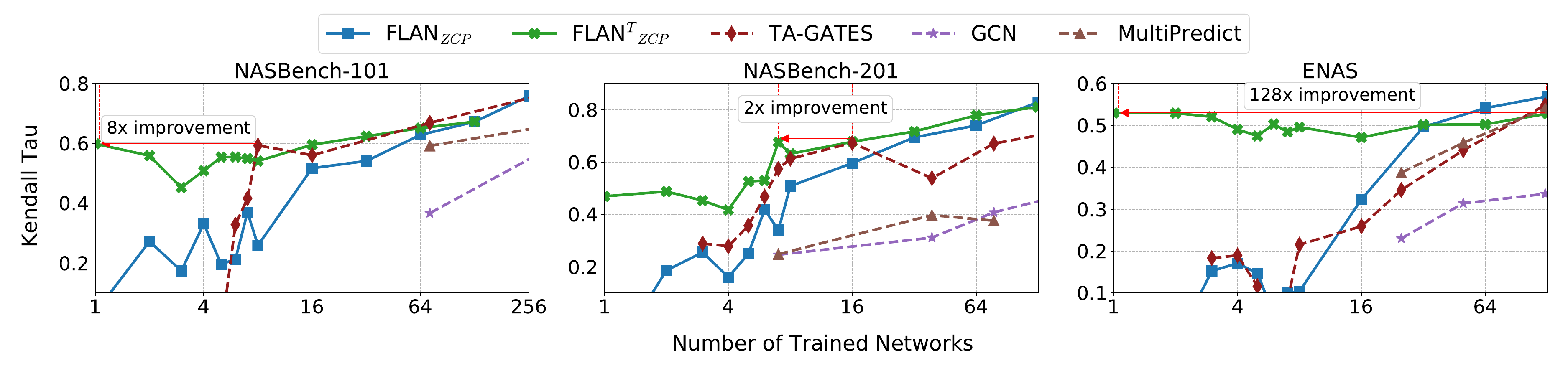}
    \caption{\GN{} sample efficiency compared to prior work. Experimental settings match \cite{tagates} and Table \ref{tab:motivate_arch}.}
    \label{fig:tagates_sampeff}
  \vspace{-5mm}
\end{figure*}

The primary components of \GN{} which significantly boost predictor performance are the residual connection in Eqn \ref{eq:dgf}, the learned operation attention mechanism $W_{o}$ in Eqn \ref{eq:dgf} and Eqn \ref{eq:graph-attention-output} and the pair-wise attention in the GAT module. These modules are ensembled in the overall network architecture, and repeated 5 times. Additionally, the NASBench-301 and Network Design Spaces (NDS) search spaces \citep{NDSPaper} provide benchmarks on large search spaces of two cell architectures, the normal and reduce cells. We train predictors on these search spaces by keeping separate DGF-GAT modules for the normal and reduce cells, and adding the aggregated outputs. 

\subsection{Operation Embeddings} 
In a NN architecture, each node or edge can be an operation such as convolution, maxpool.
GNNs generally identify these operations with a one-hot vector as an attribute. 
However, different operations have widely different characteristics. To model this, TA-GATES~\citep{tagates} operation embedding tables that can be updated independently from predictor training.
Figure \ref{fig:our_network_arch} depicts the concept of an iterative operation embedding update in more detail. 
Before producing an accuracy prediction, there are $T$ time-steps (iterations) in which the operation embeddings are updated and refined. 
In each iteration, the output of GNN flow is passed to a `Backward GNN Flow' module, which performs a backward pass using a transposed adjacency matrix. 
The output of this backward pass, along with the encoding is provided to a learnable transform that provides an update to the operation embedding table. 
This iterative refinement, conducted over specified time steps, ensures that the encodings capture more information about diverse operations within the network. Refer to \ref{subsec:archdesignabl} for a detailed ablation study focusing on vital aspects of the network design.

\subsection{FLAN Encodings}
\label{subsec:flan_enc_disc}

\textbf{Supplemental Encodings: } Supervised learned encodings are representations formed by accessing accuracies of NN architectures. 
While structural, score-based and unsupervised learned encodings do not carry information about accuracy, they can still be used to distinguish between NN architectures. 
For instance, CATE~\citep{cate} learns latent representations by computational clustering, thus providing CATE may contextualize the computational characteristics of the architecture. 
ZCP provides architectural-level information by serving as proxies for accuracy. 
Consequently, supplemental encodings can optionally be fed into the MLP prediction head after the node aggregation as shown in Figure \ref{fig:our_network_arch}. 
We find that using architecture-level ZCPs can significantly improve the sample-efficiency of predictors. 
CAZ refers to the encoding resulting from the concatenation of CATE, Arch2Vec and ZCP.

\textbf{Unified Encodings and Transferring Predictors: } Transferring knowledge between different search spaces can enhance the sample efficiency of predictors. 
However, achieving this is challenging due to the unique operations and macro structures inherent to each search space. 
To facilitate cross-search-space prediction, a unified operation space is crucial. 
Our methodology is straightforward; we concatenate a unique search space index to \textit{each} operation, creating distinctive operation vectors. 
These vectors can either be directly utilized by predictors as operation embeddings or be uniquely indexed by the operation embedding table. Note that the training time for \GN{} is less than 10 minutes on a single GPU, it is thus straightforward to regenerate an indexing that supports more spaces, and re-train the predictor.
It is noteworthy that ZCPs \textit{inherently} function as unified encodings by measuring broad architectural properties of a neural network (NN). 
Conversely, Arch2Vec and CATE are cell-based encoders. 
To accommodate this, we developed new encodings for Arch2Vec and CATE within a combined search space of 1.5 million NN architectures from all our NAS benchmarks. 
\textbf{We provide predictor training and NAS results for all neural networks available on the NAS benchmark for 13 NAS} spaces in Sections \ref{subsec:all_spaces_predictor_abl} \& \ref{subsec:all_spaces_nas_abl}, and provide a sub-set of these results in the experiments section to compare fairly to related work.
To realize such a transfer of predictors from one search space to another, a predictor, initially trained on the source search space, is adapted using the unified operation encodings and subsequently retrained on the target design space. This is denoted by a T superscript (\GN{}$^{T}$) in our experiments section.

\section{Experiments}

We investigate the efficacy of different encodings of neural networks on 13 search spaces, including NB101, NB201, NB301, 9 search spaces from NDS and TransNASBench-101 Micro. Aside from our encodings investigation, we study the transferrability of predictors across NAS search spaces \textbf{(SS)}, different tasks \textbf{(T)} on TransNASBench-101 Micro, and across different datasets \textbf{(D)}: CIFAR-10 to ImageNet. 
All of our experiments follow the Best Practices for NAS checklist \cite{lindauer2019best}, detailed in the Appendix \ref{subsec:nasbestprac}. Contrary to prior work, we generate encodings for all architectures from the NAS Benchmark for evaluation. 
To effectively evaluate encodings on these NAS spaces, we generate and open source the CATE, Arch2Vec and Adjacency representations for 1487731 NN architectures. 
NAS-Bench-Suite-Zero \cite{krishnakumar2022nasbenchsuitezero} introduces a data-set of 13 zero cost proxies across 28 tasks, totalling 44798 architectures. We generate 13 zero-cost proxies on an additional \textbf{487731} (10$\times$) NN architectures to facilitate thorough experimentation of different encodings. 
Building on previous studies, we adopt Kendall-Tau (Kendall-$\tau$) rank correlation coefficient relative to ground-truth accuracy as the primary measure of predictive ability. We use a pairwise hinge ranking loss to train our predictors \cite{tagates}. 
Different NN encodings are input to a 3-layer MLP prediction head with ReLU nonlinearities, except for the output layer which has no nonlinearities. 


\textbf{NN Encodings Study:} 
Table \ref{tab:motivate_arch} provides a comparative evaluation of different encoder categories. We evaluate predictor performance on a subset of each NAS space, specifically to align with the experimental setup of prior work \citep{tagates}, and to compare fairly to it.
In Table~\ref{tab:motivate_arch}, we train encoders on a fraction of the data, such as 1\% of 7290 (72 architectures) for NB101, and then test on all 7290 test sample architectures for NB101. 
Note, however, that all other tables, unless explicitly mentioned, evaluate all NNs available in the NAS space, ensuring a more consistent and thorough evaluation approach.
Our results in Table~\ref{tab:motivate_arch} show that Supervised Learned encodings perform best, especially when supplemented with additional encodings.
Our best predictor, \GN$_{ZCP}$, delivers up-to a 15\% improvement in Kendall-$\tau$ correlation compared to the best previous result from TA-GATES. 
The results highlight the efficacy of Supervised Learned encodings, and the importance of GNN enhancements such as residual connections and the dual graph flow mechanisms introduced in FLAN. This sets a solid predictor baseline for our cross-domain transfer study.

\textbf{Cross-Domain Transfer:} The ranking quality for predictors can be very low when \textit{training from scratch} with very few samples. This is because fewer samples are typically not sufficiently representative to train a generalizable predictor. To address this, and to enable few-shot accuracy predictors, we investigate the transfer of our baseline predictor \GN{} \textbf{across NAS spaces (SS), data-sets (D), and tasks (T)}. 
We compare \GN{}$^{T}$ (\GN{} Transfer in Cross-NAS Space \textbf{(SS)} setting) with prior train-from-scratch predictors in Figure \ref{fig:tagates_sampeff}, demonstrating an order of magnitude improvement in sample efficiency.
We conduct the rest of our experiments to more comprehensively test predictor performance on the entire NAS space after training on the number of samples specified in each table. 
Compared to prior work that only tested predictors on a subset of the NAS search space~\cite{tagates, gates}, our experimental setting is more challenging, but also more comprehensively tests the generalization of our predictors. 
%
Tables~\ref{tab:cross_t_main_results}, \ref{tab:cross_d_main_result}, and \ref{tab:cross_ss_main_result}  demonstrate significantly more efficient NAS accuracy predictors resulting from our cross-domain transfer experiments when compared to predictors trained from scratch. We train the base predictor on 1024 samples on the source domain, and test the sample efficiency of \GN{}$^{T}$ on the target domain.

\begin{table}[!t]
  \centering
  
  \begin{subtable}
    \centering
    \resizebox{\linewidth}{!}{%
    \begin{tabular}{lccccHcc}
    \toprule
    \multirow{2}{*}{\makecell{\textbf{TB101 Target}\\\textbf{Task (T) Samples}}} & \multicolumn{2}{c}{\textbf{Scratch}} &  \multicolumn{5}{c}{\textbf{Transfer From TB101 Class Scene}}  \\\cmidrule(lr){2-3} \cmidrule(lr){4-8}\addlinespace[0.5ex]
                              &      16   & 128      & 0$^{*}$     &        4  &        6  &        8  &      16   \\\cmidrule(lr){1-1}\cmidrule(lr){2-3} \cmidrule(lr){4-8}\addlinespace[0.5ex]
    \textbf{AutoEncoder}      & $0.456$   &  $0.624$ & \bR$0.836$  &  $0.794$  &\bB$0.812$&  $0.799$  &  $0.808$\\
    \textbf{Class Object}     & $0.404$   &  $0.656$ & \bR$0.844$  &   $0.754$ &  $0.796$ & \bB$0.811$&  $0.799$ \\
    \textbf{Jigsaw}           & $0.350$   &  $0.608$ & \bR$0.833$  &\bB$0.821$ &   $0.809$&   $0.778$ &  $0.793$ \\
    \textbf{Room Layout}      & $0.391$   &  $0.757$ & \bR$0.831$  & \bB$0.815$&   $0.811$&  $0.811$  &  $0.808$ \\
    \textbf{Segment Semantic} & $0.644$   &\bB$0.802$& \bR$0.829$  &   $0.789$ &   $0.766$&  $0.788$  &  $0.798$ \\
    \bottomrule
    \end{tabular}
          }
    \caption{Cross Task \textbf{(T)} Transfer.}
    \label{tab:cross_t_main_results}
  \end{subtable}

  \vspace{0.5em} 

  \begin{subtable}
    \centering
    \resizebox{\linewidth}{!}{%
    \begin{tabular}{lcccccc}
    \toprule
       \multirow{2}{*}{\makecell{\textbf{NDS ImageNet (D)}\\\textbf{Samples}}} & \multicolumn{2}{c}{\textbf{Scratch}} & \multicolumn{4}{c}{\textbf{Transfer from NDS CIFAR-10}}  \\ \cmidrule(lr){2-3} \cmidrule(lr){4-7}\addlinespace[0.5ex]
      &      16 & 128   &      0$^{*}$     &        4  &        8  &        16 \\\cmidrule(lr){1-1}\cmidrule(lr){2-3} \cmidrule(lr){4-7}\addlinespace[0.5ex]
    \textbf{Amoeba}     & $0.067$ & $0.403$ &   \bR$0.660$&  $0.598$  &    $0.629$  &   \bB$0.642$ \\
    \textbf{DARTS}      & $0.063$ & $0.488$ &    $0.592$  &   $0.604$ &  \bB$0.632$    &\bR$0.664$   \\
    \textbf{ENAS}       & $0.079$ & $0.447$ & \bB$0.567$     &  $0.550$  & $0.550$     &\bR$0.569$   \\
    \textbf{NASNet}     & $0.107$ & $0.395$ & $0.394$     & $0.396$   & \bB$0.399$     &\bR$0.437$  \\
    \textbf{PNAS}       & $0.104$ & \bB$0.426$ &   $0.378$   &   $0.370$ & $0.376$     &  \bR$0.451$   \\
    \bottomrule
    \end{tabular}
          }
    \caption{Cross data-set \textbf{(D)} Transfer}
  \label{tab:cross_d_main_result}
  \end{subtable}
  
  \vspace{0.5em} 

  \begin{subtable}
    \centering
    \resizebox{\linewidth}{!}{%
    \begin{tabular}{llHcHHcccHcc}
    \toprule
    \multicolumn{2}{c}{\textbf{Search Space}} & \multicolumn{5}{c}{\textbf{Scratch}} &  \multicolumn{5}{c}{\textbf{Transfer}}  \\\cmidrule(lr){3-7} \cmidrule(lr){8-12}\addlinespace[0.5ex]
    \textbf{Source} & \textbf{Target} &      8   &      16   & 32          & 64          & 128        & 0$^{*}$    &    4     &    6     &     8    &      16   \\\cmidrule(lr){1-2}\cmidrule(lr){3-7} \cmidrule(lr){8-12}\addlinespace[0.5ex]
    \textbf{ENAS} & \textbf{Amoeba}  & $0.025$  & $0.058$   & $0.225$     & $0.293$     & $0.435$    & \bB$0.458$  & $0.421$  &     $0$     & $0.419$  & \bR$0.470$  \\
    \textbf{ENAS} & \textbf{DARTS}   & $0.038$  & $0.081$   & $0.240$     & $0.293$     & $0.514$    & \bB$0.551$  & $0.453$  &     $0$     & $0.481$  & \bR$0.567$  \\
    \textbf{DARTS} & \textbf{ENAS}    & $0.003$  & $0.099$   & $0.352$     & $0.318$     & $0.449$    & \bB$0.465$  & $0.425$  &     $0$     & $0.426$  & \bR$0.484$ \\
    \textbf{PNAS} & \textbf{NASNet}  & $0.090$  & $0.120$   & $0.216$     & $0.299$     & \bR$0.402$    &  $0.334$ & $0.227$  &     $0$     & $0.301$  & \bB$0.344$ \\
    \textbf{NASNet} & \textbf{PNAS}    & $0.035$  & $0.102$   & $0.122$     & $0.255$     & \bR$0.431$    & $0.412$  & $0.322$  &     $0$     & $0.376$  & \bB$0.430$  \\
    \bottomrule 
    \multicolumn{12}{l}{$^{*}$ 0 samples denotes the use of the pre-trained predictor without any fine-tuning} \\
    \multicolumn{12}{l}{  on the target search space.} \\
    \end{tabular}
          }
    \caption{Cross NAS Space \textbf{(SS)} Transfer}
    \label{tab:cross_ss_main_result}
  \end{subtable}
  
\end{table}

Table \ref{tab:cross_t_main_results} compares the Kendall-$\tau$ metric when \GN{} is trained from scratch, versus when it is transferred from TB101 class scene task. A pre-trained predictor on the TB101 Class Scene task outperforms training from scratch on all target tasks, with 16--128 samples, even in the absence of fine-tuning. 
Surprisingly, adding few-shot fine-tuning might degrade predictor performance.
This emphasizes that few samples on any space may not be sufficiently representative. 
For TB101, different tasks are highly-correlated (up to 0.87 Spearman-$\rho$), which may indicate that the base predictor trained on Class-Scene is sufficiently representative for other tasks as well.
Further, we investigate \textbf{cross data-set (D)} transfer in Table \ref{tab:cross_d_main_result}. 
Training a base predictor on the CIFAR-10 data-set and performing few-shot transfer learning to the NDS ImageNet dataset improves prediction accuracy substantially as the transfer dataset size increases. 

Table \ref{tab:cross_ss_main_result} studies \textbf{cross NAS search space (SS)} transfer. 
In this more challenging setting, we use our \textit{unified encodings} (Section~\ref{subsec:flan_enc_disc}) to be able to adapt a predictor from one search space to another.
Training from scratch on the target search space with 16 samples is never enough to push predictor performance beyond 0.12 KDT, whereas transfer learning from an existing search space is effective in boosting prediction accuracy both in the zero-shot case, and with as few as 4-16 samples---at least an 8$\times$ sample efficiency improvement when compared to from-scratch predictors with 128 samples.
This can provide a concrete way to reuse NAS searches, even across different search spaces and holds promise to make NAS more efficient and sustainable computationally.

\begin{table}[t!]
      \centering
      \resizebox{\linewidth}{!}{%
            \begin{tabular}{ccccccc}
            \toprule
            \multirow{2}{*}{\textbf{Transfer}} & \multirow{2}{*}{\textbf{GENNAPE}} & \multicolumn{5}{c}{\textbf{\GN{}}$^{T}$} \\ 
             & & - & \textbf{CATE} & \textbf{Arch2Vec} & \textbf{ZCP} & \textbf{CAZ}  \\ \cmidrule(lr){1-1}\cmidrule(lr){2-2}\cmidrule(lr){3-7} \addlinespace[0.5ex]
            \textbf{Zero-Shot}  & \bR$0.815$  & \bB$0.744$  & 0.710  & 0.661  & 0.702  &  0.679 \\ 
            \textbf{50 Samples}          & 0.910  & 0.930  & 0.942  & 0.936  & \bR$0.944$  & \bB$0.934$  \\ 
            \bottomrule
            \end{tabular}
      }
      \captionof{table}{Comparing to GENNAPE~\cite{gennape} in two scenarios: using a predictor pre-trained on 50k NB101 NNs directly on NB201 without fine-tuning, and transferring the same predictor to NB201 with 50 NN accuracies. Avg. Spearman-$\rho$ over 5 trials is reported. Note that Spearman-$\rho$ is used in this experiment instead of KDT to be able to compare to GENNAPE.}
      \label{tab:gennape_nb201_compare}
\end{table}

\begin{table}[t!]
\vspace{-3mm}
    \centering
      \resizebox{0.9\linewidth}{!}{%

            \begin{tabular}{lcccccc}
            \toprule
                           
            &       \multirow{2}{*}{\textbf{CDP}}    & \multicolumn{5}{c}{\textbf{\GN{}}$^{T}$}   \\
            &                                        & -        & \textbf{CATE} & \textbf{Arch2Vec} & \textbf{ZCP} & \textbf{CAZ}  \\ \cmidrule(lr){2-2}\cmidrule(lr){3-7} \addlinespace[0.5ex]
            \textbf{Samples} & 100                   &   16     &     16        &      16           &  16          & 16             \\
            \textbf{Kendall-$\tau$}    & 0.531                 &  $0.567$ &  $0.528$      &    $0.565$        &  \bR$0.622$  & \bB$0.620$      \\
            \bottomrule
            \end{tabular}
      }
      \captionof{table}{Comparing CDP \cite{cdp_crossdomain} with our predictor (\GN{}$^{T}$) and supplementary encoding variants on DARTS. Results show the average Kendall-$\tau$  over 5 trials.}
      \label{tab:cdp_comparison}
\end{table}

%



\begin{figure*}[t!]
    \vspace{-.3cm}
    \centering
    \includegraphics[width=1.97\columnwidth]{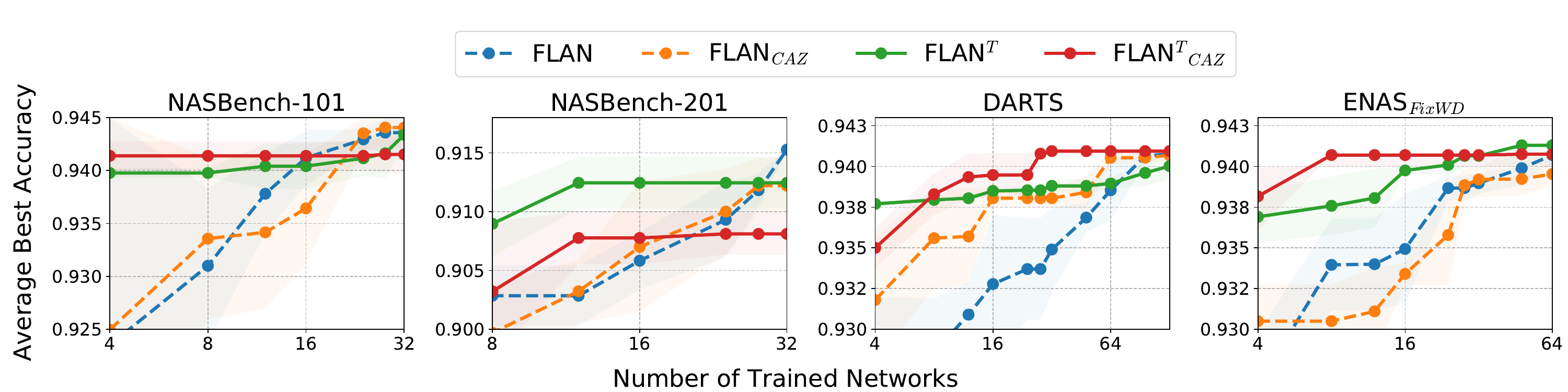}
    \caption{End-to-end NAS using an iterative sampling algorithm. \GN{}$^{T}$ improves performance for low sample counts.}
    \label{fig:nas_results}
\end{figure*}

\textbf{Comparisons to Prior Cross-Domain Transfer:} 
To compare fairly to prior work, we replicate the experimental settings of GENNAPE~\citep{gennape} and CDP~\citep{cdp_crossdomain} in Tables \ref{tab:gennape_nb201_compare} and \ref{tab:cdp_comparison} respectively.
\begin{table*}[t!]
  \centering
  \resizebox{0.96\linewidth}{!}{%
\begin{tabular}{lcccccHHcccccHcHccc}
\toprule
  & \multicolumn{15}{c}{\textbf{NASBench-101}} & \multicolumn{3}{c}{\textbf{NASBench-201}} \\ \cmidrule(lr){2-16}\cmidrule(lr){17-19} \addlinespace[0.5ex]
  
  &  \multirow{2}{*}{\textbf{BONAS}} & \textbf{Aging}     & \multirow{2}{*}{\textbf{BRPNAS}} & \multicolumn{2}{c}{\textbf{Zero-Cost NAS (W)}} & \multirow{2}{*}{\textbf{GENNAPE}} & \multicolumn{4}{c}{\multirow{2}{*}{\textbf{\GN}$_{CAZ}$}} &  \multicolumn{5}{c}{\multirow{2}{*}{\textbf{\GN}$^{T}_{CAZ}$}}     & \multirow{2}{*}{\textbf{GENNAPE}}  & \multicolumn{2}{c}{\multirow{2}{*}{\textbf{\GN}$^{T}_{CAZ}$}} \\ 
  
  &                                  & \textbf{Evo. (AE)} &                                  & \textbf{AE (15k)} & \textbf{RAND (3k)}         &         & & & &                                                     & & & & &                                                            &          &          &                \\ \cmidrule(lr){1-1}\cmidrule(lr){2-2}\cmidrule(lr){3-3}\cmidrule(lr){4-4}\cmidrule(lr){5-6}\cmidrule(lr){7-7}\cmidrule(lr){8-11}\cmidrule(lr){12-16}\cmidrule(lr){17-17}\cmidrule(lr){18-19}\addlinespace[0.5ex]
  
  \textbf{Trained models} &  1000    & 418                & 140                              & 50                & 34                         & 50000     & $0$     & $8$     & {$16$}    &  \bR{$50$}       & \bB{$0$}     & {$8$}     & $19$    &  $50$   & \bR$128$   & \bB$50$       & \bR$32$ & $64$                       \\
  \textbf{Test Acc. [\%]} &  94.22   & 94.22              & 94.22                            & 94.22             & 94.22                      &\bR$95.05$ & $91.08$ &  $93.58$ & {$94.22$} & \bR{$94.84$}     & \bB{$94.16$} & {$94.16$} & $94.22$ & $94.34$ & \bB$95.04$ & 93.27    & \bB$93.30$ &      \bR$93.73$               \\
  
\bottomrule
  \end{tabular}%
    }
\caption{A study on the number of trained models required to achieve a specified test accuracy.}
\label{tab:nas1_comp}

  \vspace{-5mm}
\end{table*}
GENNAPE performs cross-search-space transfer using a base predictor trained on 50k NN architectures on NB101, transferred to NB201. 
Table~\ref{tab:gennape_nb201_compare} shows that GENNAPE is more effective at zero-shot transfer but lags behind FLAN with 50 samples of transfer learning.
Note that GENNAPE is an ensemble model using a weighted average of multiple predictors as well as two pairwise classifiers. While the zero-shot performance of this ensemble performs well, all variants of our \GN{} with supplementary encodings outperforms the GENNAPE predictor ensemble after fine-tuning. 

\begin{table}[t]
  \centering
  \resizebox{\linewidth}{!}{%
  \begin{tabular}{ccccccc}
\toprule
& \multicolumn{5}{c}{\textbf{Cross-Transfer}}\\
 & \textbf{D}& \textbf{T} & \textbf{SS} & \textbf{SS+D} & \textbf{SS+T} & \textbf{SS+D+T} \\ \cmidrule(lr){1-1} \cmidrule(lr){2-7} \addlinespace[0.5ex]
\textbf{Kendall-}$\tau$  &  \textcolor{darkgreen}{\textbf{(+)} 0.47} & \textcolor{darkgreen}{\textbf{(+)} 0.32} & \textcolor{darkgreen}{\textbf{(+)} 0.28} & \textcolor{darkgreen}{\textbf{(+)} 0.17}  &  \textcolor{darkgreen}{\textbf{(+)} 0.16} & \textcolor{darkgreen}{\textbf{(+)} 0.10}\\
\bottomrule
\end{tabular}%
}
\caption{Enhancement in Kendall-$\tau$ when transferring predictor across combinations of NAS search spaces \textbf{(SS)}, task \textbf{(T)} and data-set \textbf{(D)} in contrast to training from scratch on the target domain from 16 samples. Reported average across suitable sub-sets of Table \ref{tab:cross_t_main_results}, \ref{tab:cross_d_main_result}, \ref{tab:cross_ss_main_result}, \ref{tab:training_from_scratch_allspaces}, \ref{tab:transfer_allspaces}.}
\label{tab:cross_all_study}
\end{table}

In Table \ref{tab:cdp_comparison}, we compare the cross search space adaptation methods LMMD + PSP~\cite{LMMD_CDP} introduced in CDP \cite{cdp_crossdomain}. CDP employs a progressive strategy, using NASBench-101, NASBench-201 and Tiny DARTS \cite{cdp_crossdomain} for cross NAS Space \textbf{(SS)} transfer, we use a single source space (ENAS) for cross NAS Space \textbf{(SS)} transfer. 
We find that our predictor with ZCP supplementary encoding and transfer offers over $5\times$ better sample efficiency and 17\% better Kendall-$\tau$ correlation. These studies further highlight the importance and effectiveness of supplementary encodings in predictor sample-efficiency.
 
Effectiveness of predictor cross-domain transfer depends not only on the predictor design and supplementary encodings, but on the nature of the source and target NAS spaces as well. We find that in some cases where there are very few samples, it may be more beneficial to pre-train a predictor. We train predictors from scratch and with transfer on all 13 NAS spaces, encompassing over 1.5 million NN architectures in the Appendix. Using this data, in Table \ref{tab:cross_all_study}, 
we summarize these many experiments by showing the average \textit{improvement} in Kendall-$\tau$ metric across combinations of \textbf{SS, D, and T} transfers.
All modes of cross-domain transfer improve upon predictors that are trained from scratch, further confirming the promise of this approach in creating transferable and reusable NAS predictors.

\vspace{2mm}


\textbf{Neural Architecture Search:} 
To gauge the NAS-efficiency of FLAN, we implement NAS search using the iterative sampling algorithm introduced by~\citet{brpnas}. 
With a budget of $n$ models per iteration and $m$ models in the search space, we use our predictor to rank the entire search space, and then select the best $\frac{n}{2}$ models. 
We sample the next $\frac{n}{2}$ models from the top $max(512, \frac{m}{2^{i}})$ models, where $i$ is the iteration counter.
Table \ref{tab:nas1_comp} compares our results to the best sample-based NAS results found in the literature. We achieve the same test accuracy as Zero-Cost NAS (W) - Rand (3k) \cite{abdelfattah2021zero} with $2.12\times$ fewer samples on end-to-end NAS with \GN$_{CAZ}$. 
%
%
GENNAPE pre-trains their base predictor on 50k samples on NB101, and transfers it to NB201. \GN{}$^{T}_{CAZ}$ pre-trains on 48$\times$ fewer NB101 samples (1024) and finds similar architectures at 36\% fewer transfer accuracy samples. Experimental set-up detailed in \ref{subsec:all_spaces_predictor_abl}. 
%
%
Further, we compare the NAS efficiency of different encoding methods in Figure \ref{fig:nas_results}. 
We find that transfer learning helps in general, with supplemental encodings (\GN$^{T}_{CAZ}$) providing the best average performance.

\vspace{-4mm}

\section{Conclusion}
We presented a comprehensive study of NN encoding methods, demonstrating their importance in enhancing the efficiency of accuracy predictors in both scenarios of training from scratch and transfer learning. 
%
%
%
Through architectural ablations (in Table \ref{tab:module_design} \& Section \ref{subsec:archdesignabl}) and supplementary encodings, we designed a state-of-the-art accuracy predictor, FLAN, that outperforms prior work by $\sim$30\%.
%
%
We used FLAN to transfer accuracy predictors \textbf{across search-spaces (SS), data-sets (D) and tasks (T)} spanning 1.5 million NNs across 13 NAS spaces, demonstrating over $8\times$ improvement in sample efficiency, and a $2.12\times$ improvement in practical NAS sample efficiency. 
We open source our code and data-sets of supplemental encodings to encourage further research on predictor design, the role of encoding in prediction based NAS and transfer learning of predictors.

\section{Impact Statement}

We study the impact of supplementary encodings and few-shot transfer on the sample efficiency of prediction-based NAS. This study is conducted on over 1.5 million NN architectures, by generating their structural, score, unsupervised and supervised learned encodings across 13 NAS spaces. Open-sourcing these encodings and the framework will have significant positive impact on NAS research and deployment by allowing effective re-use of knowledge across neural design spaces. We demonstrate the effectiveness of few-shot transfer of predictors, significantly enhancing their predictive ability with very few accuracy samples (over $8 \times$ improvement in sample efficiency). For this, we place a strong focus on using existing NAS benchmarks, saving significantly on associated model training costs. We thoroughly investigate the effectiveness of predictor transfer on cross-task, cross-dataset and cross-NAS space scenarios (as shown in Table \ref{tab:cross_all_study}). This investigation reinforces the findings of prior studies \cite{gennape, cdp_crossdomain, multipredict}, suggesting the viability of few-shot predictor transfer across markedly different NAS landscapes. We also bring the attention of the community to supplementary encodings, which are relatively cheap to generate, and can provide a 15\% improvement in sample efficiency. NAS generally requires the training of several models during the search stage, to serve as feedback on which architectural features aid accuracy. With these findings, our paper suggests the use of benchmarks to pre-train accuracy predictors, significantly improving the sample efficiency of predictor-based NAS on downstream tasks. Our demonstration of an order of magnitude improvement in sample efficiency of predicting NN accuracy can have a positive societal impact, by drastically reducing the carbon cost of NAS. We will also open-source our framework and encodings to encourage future research on sample-efficient NAS.

\bibliography{example_paper}

\begin{thebibliography}{41}
\providecommand{\natexlab}[1]{#1}
\providecommand{\url}[1]{\texttt{#1}}
\expandafter\ifx\csname urlstyle\endcsname\relax
  \providecommand{\doi}[1]{doi: #1}\else
  \providecommand{\doi}{doi: \begingroup \urlstyle{rm}\Url}\fi

\bibitem[Abdelfattah et~al.(2021)Abdelfattah, Mehrotra, Dudziak, and Lane]{abdelfattah2021zero}
Abdelfattah, M.~S., Mehrotra, A., Dudziak, {\L}., and Lane, N.~D.
\newblock Zero-cost proxies for lightweight nas.
\newblock \emph{arXiv preprint arXiv:2101.08134}, 2021.

\bibitem[Akhauri \& Abdelfattah(2023)Akhauri and Abdelfattah]{multipredict}
Akhauri, Y. and Abdelfattah, M.~S.
\newblock Multi-predict: Few shot predictors for efficient neural architecture search, 2023.

\bibitem[Chau et~al.(2022)Chau, Dudziak, Wen, Lane, and Abdelfattah]{blox}
Chau, T. C.~P., Dudziak, {\L}., Wen, H., Lane, N.~D., and Abdelfattah, M.~S.
\newblock {BLOX}: Macro neural architecture search benchmark and algorithms.
\newblock In \emph{Thirty-sixth Conference on Neural Information Processing Systems Datasets and Benchmarks Track}, 2022.
\newblock URL \url{https://openreview.net/forum?id=IIbJ9m5G73t}.

\bibitem[Dong \& Yang(2020)Dong and Yang]{dong2020nasbench201}
Dong, X. and Yang, Y.
\newblock Nas-bench-201: Extending the scope of reproducible neural architecture search.
\newblock In \emph{International Conference on Learning Representations (ICLR)}, 2020.
\newblock URL \url{https://openreview.net/forum?id=HJxyZkBKDr}.

\bibitem[Duan et~al.(2021)Duan, Chen, Xu, Chen, Liang, Zhang, and Li]{transnasbench}
Duan, Y., Chen, X., Xu, H., Chen, Z., Liang, X., Zhang, T., and Li, Z.
\newblock Transnas-bench-101: Improving transferability and generalizability of cross-task neural architecture search.
\newblock In \emph{Proceedings of the IEEE/CVF Conference on Computer Vision and Pattern Recognition}, pp.\  5251--5260, 2021.

\bibitem[Dudziak et~al.(2020)Dudziak, Chau, Abdelfattah, Lee, Kim, and Lane]{brpnas}
Dudziak, L., Chau, T., Abdelfattah, M., Lee, R., Kim, H., and Lane, N.
\newblock Brp-nas: Prediction-based nas using gcns.
\newblock volume~33, pp.\  10480--10490, 2020.

\bibitem[Guo et~al.(2019)Guo, Zheng, Tan, Chen, Chen, Zhao, and Huang]{nat_gcn}
Guo, Y., Zheng, Y., Tan, M., Chen, Q., Chen, J., Zhao, P., and Huang, J.
\newblock \emph{NAT: Neural Architecture Transformer for Accurate and Compact Architectures}.
\newblock Curran Associates Inc., Red Hook, NY, USA, 2019.

\bibitem[Kipf \& Welling(2017)Kipf and Welling]{kipf2017semisupervised_gcnarch}
Kipf, T.~N. and Welling, M.
\newblock Semi-supervised classification with graph convolutional networks.
\newblock In \emph{International Conference on Learning Representations}, 2017.
\newblock URL \url{https://openreview.net/forum?id=SJU4ayYgl}.

\bibitem[Krishnakumar et~al.(2022{\natexlab{a}})Krishnakumar, White, Zela, Tu, Safari, and Hutter]{krishnakumar2022nasbenchsuitezero}
Krishnakumar, A., White, C., Zela, A., Tu, R., Safari, M., and Hutter, F.
\newblock Nas-bench-suite-zero: Accelerating research on zero cost proxies, 2022{\natexlab{a}}.

\bibitem[Krishnakumar et~al.(2022{\natexlab{b}})Krishnakumar, White, Zela, Tu, Safari, and Hutter]{nasbenchsuitezero}
Krishnakumar, A., White, C., Zela, A., Tu, R., Safari, M., and Hutter, F.
\newblock Nas-bench-suite-zero: Accelerating research on zero cost proxies.
\newblock In \emph{Thirty-sixth Conference on Neural Information Processing Systems Datasets and Benchmarks Track}, 2022{\natexlab{b}}.

\bibitem[Lee et~al.(2021)Lee, Lee, Chong, and Hwang]{help}
Lee, H., Lee, S., Chong, S., and Hwang, S.~J.
\newblock Help: Hardware-adaptive efficient latency prediction for nas via meta-learning.
\newblock In \emph{35th Conference on Neural Information Processing Systems (NeurIPS) 2021}. Conference on Neural Information Processing Systems (NeurIPS), 2021.

\bibitem[Lee et~al.()Lee, Ajanthan, and Torr]{snip}
Lee, N., Ajanthan, T., and Torr, P.
\newblock Snip: Single-shot network pruning based on connection sensitivity.
\newblock In \emph{International Conference on Learning Representations}.

\bibitem[Li \& Talwalkar(2019)Li and Talwalkar]{randomnas}
Li, L. and Talwalkar, A.
\newblock Random search and reproducibility for neural architecture search.
\newblock \emph{arXiv preprint arXiv:1902.07638}, 2019.

\bibitem[Lindauer \& Hutter(2019)Lindauer and Hutter]{lindauer2019best}
Lindauer, M. and Hutter, F.
\newblock Best practices for scientific research on neural architecture search.
\newblock \emph{arXiv preprint arXiv:1909.02453}, 2019.

\bibitem[Liu et~al.(2018)Liu, Zoph, Neumann, Shlens, Hua, Li, Fei-Fei, Yuille, Huang, and Murphy]{progressiveNAS}
Liu, C., Zoph, B., Neumann, M., Shlens, J., Hua, W., Li, L.-J., Fei-Fei, L., Yuille, A., Huang, J., and Murphy, K.
\newblock Progressive neural architecture search.
\newblock In \emph{Proceedings of the European conference on computer vision (ECCV)}, pp.\  19--34, 2018.

\bibitem[Liu et~al.(2019)Liu, Simonyan, and Yang]{liu2018darts}
Liu, H., Simonyan, K., and Yang, Y.
\newblock {DARTS}: Differentiable architecture search.
\newblock In \emph{International Conference on Learning Representations}, 2019.
\newblock URL \url{https://openreview.net/forum?id=S1eYHoC5FX}.

\bibitem[Liu et~al.(2022)Liu, Tang, Lv, Wang, and Sun]{cdp_crossdomain}
Liu, Y., Tang, Y., Lv, Z., Wang, Y., and Sun, Y.
\newblock Bridge the gap between architecture spaces via a cross-domain predictor.
\newblock In Oh, A.~H., Agarwal, A., Belgrave, D., and Cho, K. (eds.), \emph{Advances in Neural Information Processing Systems}, 2022.
\newblock URL \url{https://openreview.net/forum?id=nE6vnoHz9--}.

\bibitem[Mehrotra et~al.(2021)Mehrotra, Ramos, Bhattacharya, Dudziak, Vipperla, Chau, Abdelfattah, Ishtiaq, and Lane]{nbasr}
Mehrotra, A., Ramos, A. G. C.~P., Bhattacharya, S., Dudziak, {\L}., Vipperla, R., Chau, T., Abdelfattah, M.~S., Ishtiaq, S., and Lane, N.~D.
\newblock {\{}NAS{\}}-bench-{\{}asr{\}}: Reproducible neural architecture search for speech recognition.
\newblock In \emph{International Conference on Learning Representations}, 2021.
\newblock URL \url{https://openreview.net/forum?id=CU0APx9LMaL}.

\bibitem[Mehta et~al.(2022)Mehta, White, Zela, Krishnakumar, Zabergja, Moradian, Safari, Yu, and Hutter]{nbsuite_naseval}
Mehta, Y., White, C., Zela, A., Krishnakumar, A., Zabergja, G., Moradian, S., Safari, M., Yu, K., and Hutter, F.
\newblock Nas-bench-suite: Nas evaluation is (now) surprisingly easy, 2022.

\bibitem[Mellor et~al.(2021)Mellor, Turner, Storkey, and Crowley]{naswot}
Mellor, J., Turner, J., Storkey, A., and Crowley, E.~J.
\newblock Neural architecture search without training.
\newblock In \emph{International Conference on Machine Learning}, pp.\  7588--7598. PMLR, 2021.

\bibitem[Mills et~al.(2022)Mills, Han, Zhang, Chudak, Mamaghani, Salameh, Lu, Jui, and Niu]{gennape}
Mills, K.~G., Han, F.~X., Zhang, J., Chudak, F., Mamaghani, A.~S., Salameh, M., Lu, W., Jui, S., and Niu, D.
\newblock Gennape: Towards generalized neural architecture performance estimators, 2022.

\bibitem[Ming~Chen et~al.(2020)Ming~Chen, Zengfeng~Huang, and Li]{gcn_residual_gcnii}
Ming~Chen, Z.~W., Zengfeng~Huang, B.~D., and Li, Y.
\newblock Simple and deep graph convolutional networks.
\newblock 2020.

\bibitem[Ning et~al.(2022)Ning, Zhou, Zhao, Zhao, Deng, Tang, Liang, Yang, and Wang]{tagates}
Ning, X., Zhou, Z., Zhao, J., Zhao, T., Deng, Y., Tang, C., Liang, S., Yang, H., and Wang, Y.
\newblock {TA}-{GATES}: An encoding scheme for neural network architectures.
\newblock In Oh, A.~H., Agarwal, A., Belgrave, D., and Cho, K. (eds.), \emph{Advances in Neural Information Processing Systems}, 2022.
\newblock URL \url{https://openreview.net/forum?id=74fJwNrBlPI}.

\bibitem[Ning et~al.(2023)Ning, Zheng, Zhou, Zhao, Yang, and Wang]{gates}
Ning, X., Zheng, Y., Zhou, Z., Zhao, T., Yang, H., and Wang, Y.
\newblock A generic graph-based neural architecture encoding scheme with multifaceted information.
\newblock \emph{IEEE Transactions on Pattern Analysis and Machine Intelligence}, 45\penalty0 (7):\penalty0 7955--7969, 2023.
\newblock \doi{10.1109/TPAMI.2022.3228604}.

\bibitem[Pham et~al.(2018)Pham, Guan, Zoph, Le, and Dean]{enas}
Pham, H., Guan, M., Zoph, B., Le, Q., and Dean, J.
\newblock Efficient neural architecture search via parameters sharing.
\newblock In Dy, J. and Krause, A. (eds.), \emph{Proceedings of the 35th International Conference on Machine Learning}, volume~80 of \emph{Proceedings of Machine Learning Research}, pp.\  4095--4104. PMLR, 10--15 Jul 2018.
\newblock URL \url{https://proceedings.mlr.press/v80/pham18a.html}.

\bibitem[Radosavovic et~al.(2019)Radosavovic, Johnson, Xie, Lo, and Doll{\'a}r]{NDSPaper}
Radosavovic, I., Johnson, J., Xie, S., Lo, W.-Y., and Doll{\'a}r, P.
\newblock On network design spaces for visual recognition.
\newblock In \emph{ICCV}, 2019.

\bibitem[Real et~al.(2019)Real, Aggarwal, Huang, and Le]{amoebanas}
Real, E., Aggarwal, A., Huang, Y., and Le, Q.~V.
\newblock Regularized evolution for image classifier architecture search.
\newblock In \emph{Proceedings of the Thirty-Third AAAI Conference on Artificial Intelligence and Thirty-First Innovative Applications of Artificial Intelligence Conference and Ninth AAAI Symposium on Educational Advances in Artificial Intelligence}, pp.\  4780--4789, 2019.

\bibitem[Shi et~al.(2020)Shi, Pi, Xu, Li, Kwok, and Zhang]{bonas}
Shi, H., Pi, R., Xu, H., Li, Z., Kwok, J., and Zhang, T.
\newblock Bridging the gap between sample-based and one-shot neural architecture search with bonas.
\newblock volume~33, pp.\  1808--1819, 2020.

\bibitem[Tanaka et~al.(2020)Tanaka, Kunin, Yamins, and Ganguli]{synflow}
Tanaka, H., Kunin, D., Yamins, D.~L., and Ganguli, S.
\newblock Pruning neural networks without any data by iteratively conserving synaptic flow.
\newblock \emph{Advances in neural information processing systems}, 33:\penalty0 6377--6389, 2020.

\bibitem[Veli{\v{c}}kovi{\'c} et~al.(2018)Veli{\v{c}}kovi{\'c}, Cucurull, Casanova, Romero, Lio, and Bengio]{gat_bengio}
Veli{\v{c}}kovi{\'c}, P., Cucurull, G., Casanova, A., Romero, A., Lio, P., and Bengio, Y.
\newblock Graph attention networks.
\newblock In \emph{International Conference on Learning Representations}, 2018.

\bibitem[White et~al.(2020)White, Neiswanger, Nolen, and Savani]{encodingstudy}
White, C., Neiswanger, W., Nolen, S., and Savani, Y.
\newblock A study on encodings for neural architecture search.
\newblock In \emph{Advances in Neural Information Processing Systems}, 2020.

\bibitem[White et~al.(2021)White, Zela, Ru, Liu, and Hutter]{whiteperfpredictornas}
White, C., Zela, A., Ru, B., Liu, Y., and Hutter, F.
\newblock How powerful are performance predictors in neural architecture search?, 2021.

\bibitem[Yan et~al.(2020)Yan, Zheng, Ao, Zeng, and Zhang]{arch2vec}
Yan, S., Zheng, Y., Ao, W., Zeng, X., and Zhang, M.
\newblock Does unsupervised architecture representation learning help neural architecture search?
\newblock In \emph{NeurIPS}, 2020.

\bibitem[Yan et~al.(2021)Yan, Song, Liu, and Zhang]{cate}
Yan, S., Song, K., Liu, F., and Zhang, M.
\newblock Cate: Computation-aware neural architecture encoding with transformers.
\newblock In \emph{ICML}, 2021.

\bibitem[Yang et~al.(2020)Yang, Esperan{\c{c}}a, and Carlucci]{yang2019evaluation}
Yang, A., Esperan{\c{c}}a, P.~M., and Carlucci, F.~M.
\newblock Nas evaluation is frustratingly hard.
\newblock 2020.

\bibitem[Ying et~al.(2019)Ying, Klein, Christiansen, Real, Murphy, and Hutter]{nasbench101}
Ying, C., Klein, A., Christiansen, E., Real, E., Murphy, K., and Hutter, F.
\newblock Nas-bench-101: Towards reproducible neural architecture search.
\newblock In \emph{International Conference on Machine Learning}, pp.\  7105--7114. PMLR, 2019.

\bibitem[Zela et~al.(2020)Zela, Siems, Zimmer, Lukasik, Keuper, and Hutter]{nasbench301}
Zela, A., Siems, J., Zimmer, L., Lukasik, J., Keuper, M., and Hutter, F.
\newblock Surrogate nas benchmarks: Going beyond the limited search spaces of tabular nas benchmarks, 2020.
\newblock URL \url{https://arxiv.org/abs/2008.09777}.

\bibitem[Zhou et~al.(2020)Zhou, Zhou, Zhang, Loy, Yi, Zhang, and Ouyang]{econas}
Zhou, D., Zhou, X., Zhang, W., Loy, C., Yi, S., Zhang, X., and Ouyang, W.
\newblock Econas: Finding proxies for economical neural architecture search.
\newblock In \emph{2020 IEEE/CVF Conference on Computer Vision and Pattern Recognition (CVPR)}, pp.\  11393--11401, Los Alamitos, CA, USA, jun 2020. IEEE Computer Society.
\newblock \doi{10.1109/CVPR42600.2020.01141}.
\newblock URL \url{https://doi.ieeecomputersociety.org/10.1109/CVPR42600.2020.01141}.

\bibitem[Zhu et~al.(2021)Zhu, Zhuang, Wang, Ke, Chen, Bian, Xiong, and He]{LMMD_CDP}
Zhu, Y., Zhuang, F., Wang, J., Ke, G., Chen, J., Bian, J., Xiong, H., and He, Q.
\newblock Deep subdomain adaptation network for image classification.
\newblock \emph{IEEE Transactions on Neural Networks and Learning Systems}, 32\penalty0 (4):\penalty0 1713--1722, 2021.
\newblock \doi{10.1109/TNNLS.2020.2988928}.

\bibitem[Zoph \& Le(2017)Zoph and Le]{rl_nas}
Zoph, B. and Le, Q.
\newblock Neural architecture search with reinforcement learning.
\newblock In \emph{International Conference on Learning Representations}, 2017.
\newblock URL \url{https://openreview.net/forum?id=r1Ue8Hcxg}.

\bibitem[Zoph et~al.(2018)Zoph, Vasudevan, Shlens, and Le]{nasnet}
Zoph, B., Vasudevan, V., Shlens, J., and Le, Q.
\newblock Learning transferable architectures for scalable image recognition.
\newblock pp.\  8697--8710, 06 2018.
\newblock \doi{10.1109/CVPR.2018.00907}.

\end{thebibliography}
\bibliographystyle{icml2024}

\newpage
\appendix
\onecolumn

\section{Appendix}
\subsection{Best practices for NAS}
\label{subsec:nasbestprac}
\citep{encodingstudy, randomnas, nasbench101, yang2019evaluation} discuss improving reproducibility and fairness in experimental comparisons for NAS. We thus address the sections released in the NAS best practices checklist by \citep{lindauer2019best}.

\begin{itemize}
    \item \textbf{Best Practice: Release Code for the Training Pipeline(s) you use: } We release code for our Predictor, CATE, Arch2Vec encoder training set-up.
    \item \textbf{Best Practice: Release Code for Your NAS Method: } We release our code publicly for the BRP-NAS style NAS search. We do not introduce a new NAS method. 
    \item \textbf{Best Practice: Use the Same NAS Benchmarks, not Just the Same Datasets: } We use the NASBench-101, NASBench-201, NASBench-301, NDS and TransNASBench-101 datasets for evaluation. We also use a sub-set of Zero Cost Proxies from NAS-Bench-Suite-Zero.
    \item \textbf{Best Practice: Run Ablation Studies: } We run ablation studies for the design of \GN{} in Table \ref{tab:flan_timesteps}, Table \ref{tab:module_design2}, Table \ref{tab:nb101_arch_design} and Table \ref{tab:nb201_arch_design}. We conduct ablation studies with different supplementary encodings in the main paper.
    \item \textbf{Best Practice: Use the Same Evaluation Protocol for the Methods Being Compared: } We use the same evaluation protocol as TAGATES when comparing encoders across literature. We provide additional larger studies that all follow the same evaluation protocol.
    \item \textbf{Best Practice: Evaluate Performance as a Function of Compute Resources: } In this paper, we study the sample efficiency of encodings. We report results in terms of the 'number of trained models required'. This directly correlates with compute resources, depending on the NAS space training procedure.
    \item \textbf{Best Practice: Compare Against Random Sampling and Random Search: } We propose a predictor - encoder design methodology, not a NAS method. We use SoTA BRP-NAS style NAS algorithm for comparing with existing literature. 
    \item \textbf{Best Practice: Perform Multiple Runs with Different Seeds: } Our appendix contains information on number of trials as well as tables for all NAS spaces with standard deviation in Table \ref{tab:training_from_scratch_allspaces} and Table \ref{tab:transfer_allspaces}.
    \item \textbf{Best Practice: Use Tabular or Surrogate Benchmarks If Possible: } All our evaluations are done on publicly available Tabular and Surrogate benchmarks.
\end{itemize}

\begin{table*}[b]
    \centering
\resizebox{0.7\linewidth}{!}{%
\centering
\begin{tabular}{@{}lcccc@{}}
\toprule
\multicolumn{1}{l}{\textbf{Search space}} & \multicolumn{1}{c}{\textbf{Tasks}} & 
\multicolumn{1}{c}{\textbf{Num.\ ZC proxies}} & 
\multicolumn{1}{c}{\textbf{Num.\ architectures}} &
\multicolumn{1}{c}{\textbf{Total ZC proxy evaluations}} \\ \cmidrule(lr){1-1} \cmidrule(lr){2-2} \cmidrule(lr){3-3} \cmidrule(lr){4-4} \cmidrule(lr){5-5} \addlinespace[0.5ex]
NAS-Bench-101 & 1 & 13 & 423\,625 & 5\,507\,125 \\
NAS-Bench-201 & 1 & 13 & 15\,625 & 203\,125 \\
DARTS                                   & 1 & 13 & 5000 & 65000 \\
ENAS                                    & 1 & 13 & 4999 & 64987 \\
PNAS                                    & 1 & 13 & 4999 & 64987 \\
NASNet                                  & 1 & 13 & 4846 & 62998 \\
AmoebaNet                               & 1 & 13 & 4983 & 64779 \\
DARTS$_{FixWD}$                         & 1 & 13 & 5000 & 65000 \\
DARTS$_{LRWD}$                          & 1 & 13 & 5000 & 65000 \\
ENAS$_{FixWD}$                          & 1 & 13 & 5000 & 65000 \\
PNAS$_{FixWD}$                          & 1 & 13 & 4559 & 59267 \\
TransNASBench-101 Micro& 7 & 12 & 4096 & 344\,064\\
\textbf{Total} & 18 & 13 & 512\,308 & 6\,631\,332 \\
\bottomrule
\end{tabular}%
}
\caption{Overview of ZC proxy evaluations in our work. ZCP for TransNASBench-101 Micro and NASBench201 are borrowed from \cite{nasbenchsuitezero}. }
\label{tab:zcp_details}
\end{table*}

\subsection{Neural Architecture Design Spaces}
\label{subsec:nasspacesabl}
In this paper, multiple distinct neural architecture design spaces are studied. Both NASBench-101\citep{nasbench101} and NASBench-201\citep{dong2020nasbench201} are search spaces based on cells, comprising 423,624 and 15,625 architectures respectively. NASBench-101 undergoes training on CIFAR-10, whereas NASBench-201 is trained on CIFAR-10, CIFAR-100, and ImageNet16-120. NASBench-301\citep{nasbench301} serves as a surrogate NAS benchmark, containing a total of $10^{18}$ architectures. TransNAS-Bench-101\citep{transnasbench} stands as a NAS benchmark that includes a micro (cell-based) search space with 4096 architectures and a macro search space embracing 3256 architectures. In our paper, we only study TransNASBench-101 Micro as that is a cell-based search space. These networks are individually trained on seven different tasks derived from the Taskonomy dataset. The NASLib framework unifies these search spaces. The NAS-Bench-Suite-Zero\citep{nasbenchsuitezero} further extends this space by incorporating two datasets from NAS-Bench-360, SVHN, and another four datasets from Taskonomy. Further, the NDS\citep{NDSPaper} spaces are described in Table \ref{tab:nas_details} borrowed from the original paper. Additionally, the NDS data-set has `$FixWD$' data-sets which indicate that the width and depth do not vary in architectures. The NDS data-set has $LRWD$ data-sets which indicate that the learning rates do not vary in architectures. We do not include learning rate related representations in our predictor, while it is possible and may benefit performance. We only look at architectural aspects of the NAS design problem.

\begin{table*}[h!]
    \centering
    \resizebox{0.6\linewidth}{!}{%
        \begin{tabular}{lcccr}
        \toprule
         & num ops & num nodes & output & num cells (B)\\  \cmidrule(lr){1-1} \cmidrule(lr){2-2} \cmidrule(lr){3-3} \cmidrule(lr){4-4} \cmidrule(lr){5-5} \addlinespace[0.5ex]
         NASNet~\citep{nasnet} & 13 & 5 & L & 71,465,842\\
         Amoeba~\citep{amoebanas} &  8 & 5 & L & 556,628\\
         PNAS~\citep{progressiveNAS}    &  8 & 5 & A & 556,628\\
         ENAS~\citep{enas}   &  5 & 5 & L & 5,063\\
         DARTS~\citep{liu2018darts}   &  8 & 4 & A & 242\\
         \bottomrule
        \end{tabular}%
        }
\caption{\textbf{NAS design spaces.} NDS \citep{NDSPaper} summarizes the cell structure for five NAS design spaces. This table lists the number of candidate ops (5$\times$5 conv, 3$\times$3 max pool), number of nodes (excluding the inputs), and which nodes are concatenated for the output (`A' if `all' nodes, `L' if `loose' nodes not used as input to other nodes). Given $o$ ops to choose from, there are $o^2{\cdot}(j{+}1)^2$ choices when adding the $j^{th}$ node, leading to $o^{2k}{\cdot}((k{+}1)!)^2$ possible cells with $k$ nodes (of course many of these cells are redundant). The spaces vary substantially; indeed, even exact candidate ops for each vary.}
\label{tab:nas_details}
\end{table*}

\subsection{Additional Results}
In this sub-section, we provide more complete versions of some of the graphs and results in the main paper.

\begin{figure}[t!]
    \centering
    \includegraphics[width=\columnwidth]{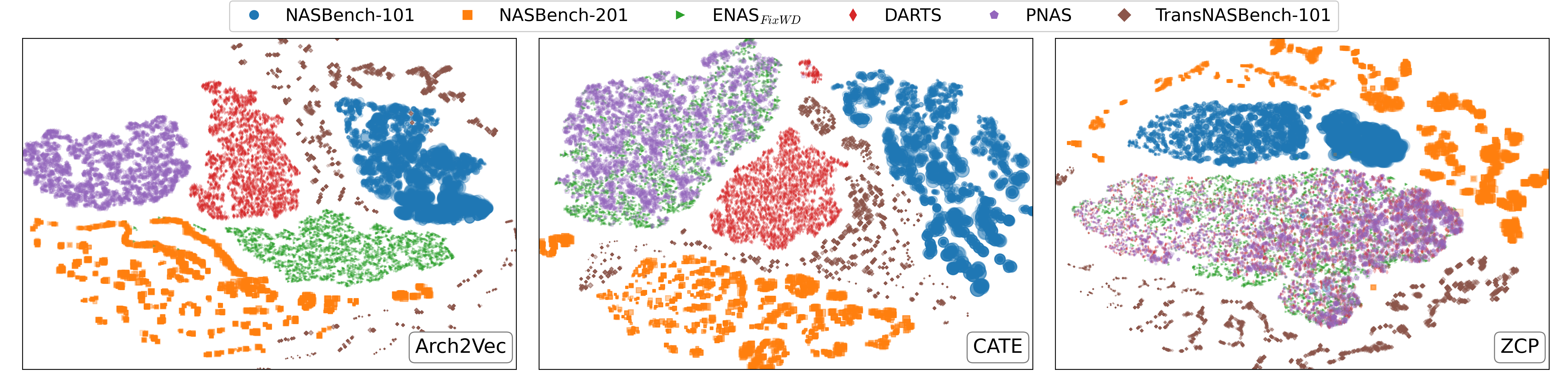}
    \caption{t-SNE scatterplot of the encodings for a set of architecture families using the ZCP, unified Arch2Vec and unified CATE encodings. Best viewed in color.}
    \label{fig:tsne_all}
\end{figure}

The t-SNE scatterplot showcased in Figure \ref{fig:tsne_all} demonstrates distinct clustering patterns associated with Arch2Vec based on different search spaces. This pattern is attributed to the binary indexing approach utilized in operations representation. Similar clustering tendencies are also observable for CATE and ZCP. However, it's noteworthy that search spaces like ENAS$_{FixWD}$ and PNAS tend to cluster more closely in the CATE representation. This proximity is influenced by the similarities in their parameter counts. In the case of ZCP, DARTS shows a tendency to cluster within the ENAS$_{FixWD}$ and PNAS spaces, which can be attributed to shared zero-cost characteristics. These observations highlight the distinct nature of the encoding methodologies employed by ZCP, Arch2Vec, and CATE. Quantitative analysis reveals the correlation of parameter count with the respective representations as 0.56 for ZCP, 0.38 for CATE, and 0.13 for Arch2Vec. This quantitative insight underscores the differential impact of encoding strategies on the parameter space representation across various search spaces.

\subsection{Neural Architecture Search on NASBench-201 CIFAR-100}

\begin{figure}[h!]
    \centering
    \begin{minipage}{.54\linewidth}
        \centering
        \resizebox{\linewidth}{!}{%
            \begin{tabular}{cccccccc}
                \toprule
                Source & NB201 & NB301 & NB101 & NB201 & PNAS$_{FixWD}$ & \multicolumn{2}{c}{ENAS$_{FixWD}$} \\
                Target & NB101 & NB201 & NB301 & TB101 & Amoeba & \multicolumn{2}{c}{PNAS$_{FixWD}$} \\ 
                \cmidrule(lr){1-1} \cmidrule(lr){2-8} 
                Source & NASNet & DARTS & ENAS & PNAS & DARTS$_{LRWD}$ & DARTS$_{FixWD}$ & PNAS \\
                Target & ENAS$_{FixWD}$ & NASNet & DARTS & ENAS & PNAS & DARTS$_{LRWD}$ & DARTS$_{FixWD}$ \\ 
                \bottomrule
            \end{tabular}
        }
        \caption{Source and Target Spaces for experiments unless specified otherwise.}
        \label{tab:all_ss_spaces_mapping}
    \end{minipage}
    \hfill
    \begin{minipage}{.45\linewidth}
        \centering
        \includegraphics[width=\linewidth]{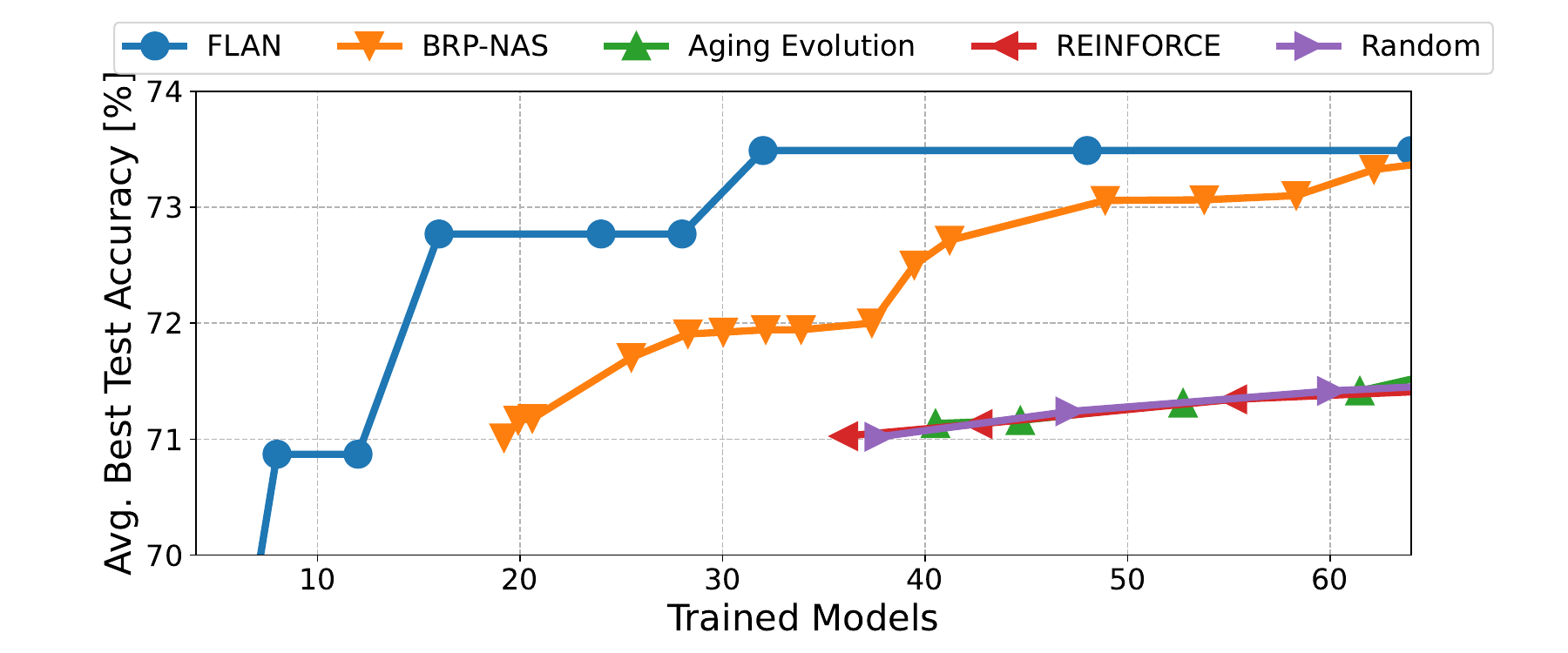}
        \caption{\GN{} with the iterative sampling search algorithm (BRP-NAS) outperforms other popular search methodologies on NASBench-201 CIFAR100.}
        \label{fig:nb2_cf100_brpnas}
    \end{minipage}
\end{figure}

To demonstrate the effectiveness of our predictor on NAS on more search spaces, we compare \GN{} with BRP-NAS to compare predictors, as well as other NAS search methodologies in Figure \ref{fig:nb2_cf100_brpnas}.

\subsection{On run-time of our predictor}

Training \GN{} is extremely efficient, with our median training time being approximately 7.5 minutes. This implies that modifications to search space descriptions or indexing can be done trivially and \GN{} can be re-trained trivially. Further, generating the unified Arch2Vec and CATE encodings can both be done in under an hour on a consumer GPU. The time to transfer to a new search space depends upon the number of samples, our maximum time for transfer in tests was approximately 1 minute. Finally, for inference during NAS, we can evaluate approximately 160 architectures per second.

\begin{figure*}[t!]
    \centering
    \includegraphics[width=0.9\linewidth]{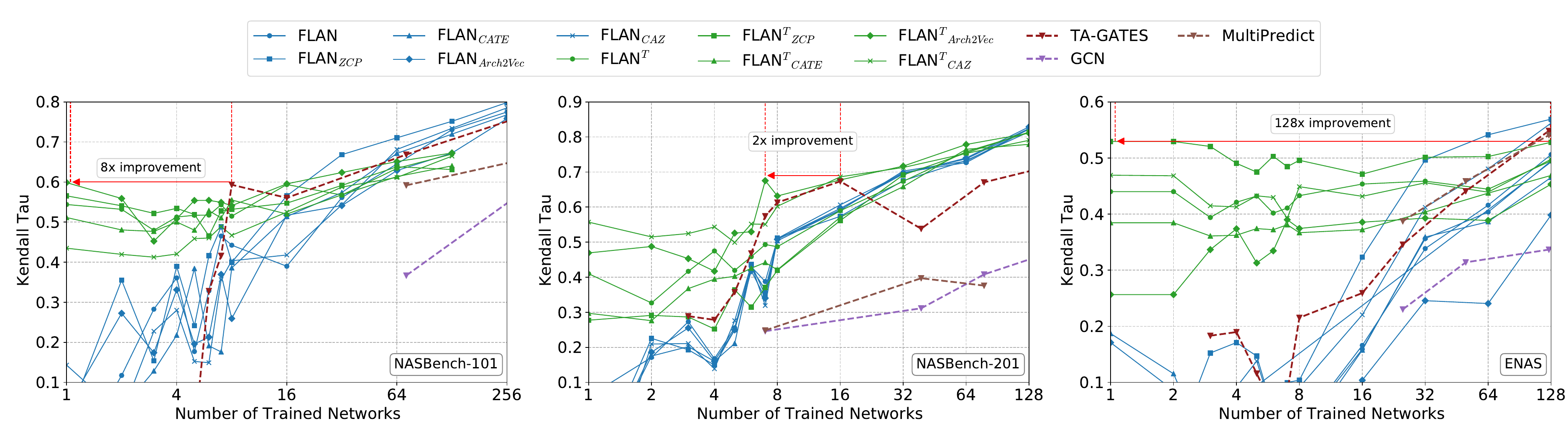}
    \caption{Prediction accuracy with different numbers of trained NNs. We investigate the impact of supplemental and unified encodings with FLAN, and compare to prior work. X-axis is logarithmic. Source space for NASBench-201 is NASBench-101 and vice versa. Source space for ENAS is DARTS.}
    \label{fig:tagates_sampeff_apndx}
\end{figure*}

\begin{figure*}[t!]
    \vspace{-.5cm}
    \centering
    \includegraphics[width=0.9\linewidth]{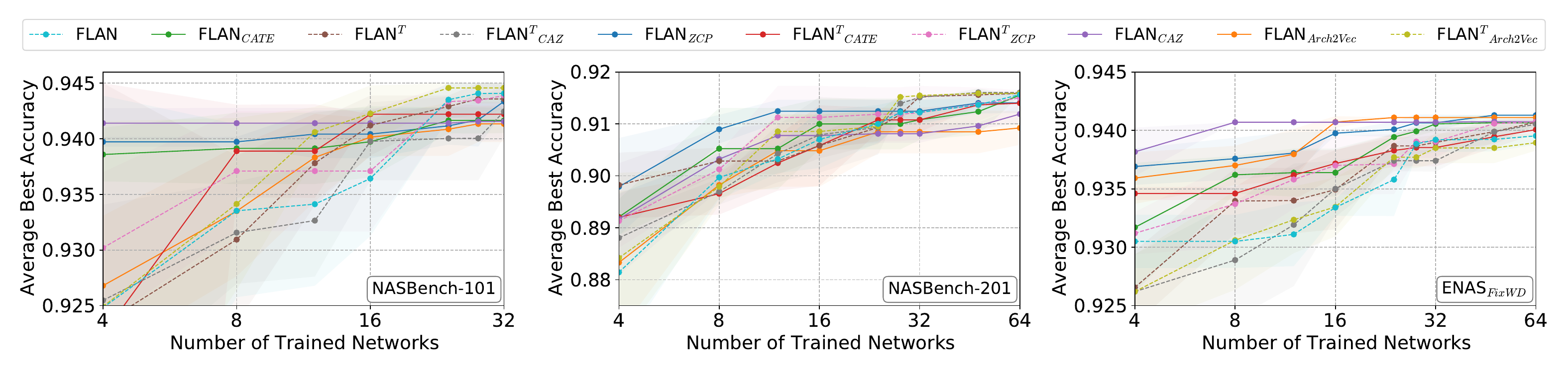}
    \caption{End-to-end NAS with different predictors using an iterative sampling search algorithm. \GN{}$^{T}$ improves search efficiency in the low sample count region. Source search-space for NASBench-201 is NASBench-101 and vice versa. Source space for ENAS$_{FixWD}$ is PNAS.}
    \label{fig:nas_results_apdx}
\end{figure*}

\subsection{Experimental Setup}
\label{subsec:expsetup_abl}
In this paper, we focus on standardizing our experiments on entire NAS Design spaces. We open source our code and generated encodings to foster further research. Additionally, we list the primary experimental hyperparameters in Table \ref{tab:hyperparameters}.

\begin{table}[h!]
    \centering
    \resizebox{\linewidth}{!}{%
\begin{tabular}{llll}
\toprule
\textbf{Hyperparameter} & \textbf{Value} & \textbf{Hyperparameter} & \textbf{Value} \\ \cmidrule(lr){1-1} \cmidrule(lr){2-2} \cmidrule(lr){3-3} \cmidrule(lr){4-4} \addlinespace[0.5ex]
Learning Rate & 0.001 & Weight Decay & 0.00001 \\
Number of Epochs & 150 & Batch Size & 8 \\
Number of Transfer Epochs & 30 & Transfer Learning Rate & 0.001 \\
Graph Type & `DGF+GAT ensemble' & Op Embedding Dim & 48 \\
Node Embedding Dim & 48 & Hidden Dim & 96 \\
GCN Dims & [128, 128, 128, 128, 128] & MLP Dims & [200, 200, 200] \\
GCN Output Conversion MLP & [128, 128] & Backward GCN Out Dims & [128, 128, 128, 128, 128] \\
OpEmb Update MLP Dims & [128] & NN Emb Dims & 128 \\
Supplementary Encoding Embedder Dims & [128, 128] & Number of Time Steps & 2 \\
Number of Trials & 9 & Loss Type & Pairwise Hinge Loss \citep{tagates} \\
\bottomrule
\end{tabular}%
}
\caption{Hyperparameters used in main table experiments.}
\label{tab:hyperparameters}
\end{table}

It is important to note that our results for the PATH encoding are generated with the naszilla hyper-paramaters described in Table \ref{tab:path_hyp}. Upon reproducing their set-up on our own MLP network architecture, adjacency representation was much better than path encoding. This further highlights the importance of predictor design. In Table \ref{tab:pathstudy}, we see that their MetaNN ourperforms our NN design, but only for the PATH encoding. 

\begin{figure}[h!]
    \centering
    \begin{minipage}{.48\linewidth}
  \centering
  \resizebox{\linewidth}{!}{%
            \begin{tabular}{lcccHHccc}
            \multicolumn{9}{c}{NASBench-101}\\ \toprule
            Timesteps & DGF & Leaky & KQV & Operation & Attention & \multicolumn{3}{c}{Number Of Samples}\\
             & Residual  & ReLU & Projection & Attention  &  Residual & 8 & 16 & 32 \\ \cmidrule(lr){1-1} \cmidrule(lr){2-2} \cmidrule(lr){3-3} \cmidrule(lr){4-4}\cmidrule(lr){5-5} \cmidrule(lr){6-6} \cmidrule(lr){7-7}\cmidrule(lr){8-8}\cmidrule(lr){9-9} \addlinespace[0.5ex]
            1 & \ding{51} & \ding{55} & \ding{51} & \ding{51} & \ding{51} & 0.3945 & 0.4939 & 0.5340\\
            1 & \ding{55} & \ding{51} & \ding{51} & \ding{51} & \ding{51} & 0.2425 & 0.4434 & 0.5448\\
            1 & \ding{51} & \ding{51} & \ding{51} & \ding{51} & \ding{51} & 0.3348 & 0.5230 & 0.5829\\
            1 & \ding{51} & \ding{51} & \ding{55} & \ding{51} & \ding{51} & 0.4129 & 0.5301 & 0.5442\\
            1 & \ding{55} & \ding{55} & \ding{51} & \ding{51} & \ding{51} & 0.3132 & 0.4311 & 0.5454\\
            1 & \ding{55} & \ding{55} & \ding{55} & \ding{51} & \ding{51} & \bR$0.4791$ & 0.4658 & 0.5123\\
            1 & \ding{55} & \ding{51} & \ding{55} & \ding{51} & \ding{51} & 0.4595 & 0.5098 & \bR$0.5904$\\
            2 & \ding{51} & \ding{55} & \ding{55} & \ding{51} & \ding{51} & 0.4628 & 0.5299 & 0.4825\\
            2 & \ding{55} & \ding{51} & \ding{51} & \ding{51} & \ding{51} & 0.4487 & 0.4832 & 0.5582\\
            2 & \ding{55} & \ding{51} & \ding{55} & \ding{51} & \ding{51} & 0.3403 & \bR$0.5428$ & 0.5495\\
            2 & \ding{55} & \ding{55} & \ding{51} & \ding{51} & \ding{51} & 0.3562 & 0.4420 & 0.4737\\
            2 & \ding{55} & \ding{55} & \ding{55} & \ding{51} & \ding{51} & 0.2640 & 0.5162 & 0.5316\\
            2 & \ding{51} & \ding{51} & \ding{51} & \ding{51} & \ding{51} & 0.3939 & 0.5081 & 0.5684\\
            3 & \ding{51} & \ding{51} & \ding{55} & \ding{51} & \ding{51} & 0.3899 & 0.4633 & 0.5446\\
            3 & \ding{51} & \ding{51} & \ding{51} & \ding{51} & \ding{51} & 0.4756 & 0.5340 & 0.5484\\
            3 & \ding{51} & \ding{55} & \ding{55} & \ding{51} & \ding{51} & 0.3020 & 0.5025 & 0.5616\\
            3 & \ding{55} & \ding{51} & \ding{55} & \ding{51} & \ding{51} & 0.2957 & 0.3765 & 0.5607\\
            3 & \ding{55} & \ding{51} & \ding{51} & \ding{51} & \ding{51} & 0.3280 & 0.4647 & 0.5552\\
            3 & \ding{55} & \ding{55} & \ding{51} & \ding{51} & \ding{51} & 0.3674 & 0.5241 & 0.5291\\\bottomrule
            \end{tabular}%
            }
            \caption{Results of architecture design ablation. Tested on 1000 randomly sampled architectures. Average over 3 trials. Results depict the Kendall-$\tau$ Correlation of \GN{} with different DGF GAT module implementations.}
            \label{tab:nb101_arch_design} 
    \end{minipage}
    \hfill
    \begin{minipage}{.48\linewidth}
        \centering
        \resizebox{0.6\linewidth}{!}{%
        \begin{tabular}{llll}
        \toprule
        Parameter & Value & Parameter & Value \\ \cmidrule(lr){1-1} \cmidrule(lr){2-2} \cmidrule(lr){3-3} \cmidrule(lr){4-4} \addlinespace[0.5ex]
        Loss & MAE & NN Depth & 10 \\
        NN Width & 20 & Epochs & 200 \\
        Batch Size & 32 & LR & 0.01 \\\bottomrule
        \end{tabular}%
        }
        \caption{Hyperparameters used to generate PATH results.}
        \label{tab:path_hyp}

        \vspace{10pt} 
        \centering
        \resizebox{0.95\linewidth}{!}{%
        \begin{tabular}{llllll}
        \toprule
        Training Samples & Adj MetaNN & Adj NN  & Path MetaNN & Path NN\\\cmidrule(lr){1-1} \cmidrule(lr){2-2} \cmidrule(lr){3-3} \cmidrule(lr){4-4}\cmidrule(lr){5-5} \addlinespace[0.5ex]
        72      & 0.057      & \bR$0.3270$  & \bR$0.3875$      & -0.0315                \\
        364     & 0.1464     & \bR$0.4647$  & \bR$0.6967$      & -0.0363         \\
        729     & 0.2269     & \bR$0.5141$  & \bR$0.7524$      & -0.0023        \\\bottomrule
        \end{tabular}%
        }
        \caption{Study on PATH Encoding for NASBench-101. Tested on 7290 samples.}
        \label{tab:pathstudy}
        \vspace{10pt} 
    \centering
    \resizebox{\linewidth}{!}{%
        \begin{tabular}{ccccc}
              \toprule
              Timesteps &      DARTS$_{FixWD}$ &       ENAS$_{FixWD}$ &              NB101 &              TB101  \\
              \cmidrule(lr){1-1} \cmidrule(lr){2-2} \cmidrule(lr){3-3} \cmidrule(lr){4-4} \cmidrule(lr){5-5} \addlinespace[0.5ex]
              1      &  \bR$0.4870_{0.0002}$ &  $0.4653_{0.0031}$ &  $0.7017_{0.0007}$ &  $0.7789_{0.0002}$ \\
              2      &  $0.4632_{0.0003}$ &  $0.4799_{0.0010}$ &  $0.7129_{0.0001}$ &  \bR$0.7939_{0.0001}$ \\
              4      &  $0.4801_{0.0001}$ &  \bR$0.4803_{0.0012}$ &  \bR$0.7133_{0.0001}$ &  $0.7907_{0.0002}$ \\
              \bottomrule
              \end{tabular}%
    }
\caption{We study the importance of time-steps in the \GN{} predictor design. 128 samples are used to train, and tested on the entire NAS space.}
\label{tab:flan_timesteps} 
    \end{minipage}
\end{figure}

\begin{table}[t!]
    \centering
    \resizebox{\linewidth}{!}{%
          \begin{tabular}{ccccccccccc}
              \toprule
              Forward & Backward        & NB101 & NB201 & NB301               & Amoeba               & PNAS                & NASNet              & DARTS$_{FixWD}$     & ENAS$_{FixWD}$      & TB101 \\
              \cmidrule(lr){1-2}\cmidrule(lr){3-3}\cmidrule(lr){4-4}\cmidrule(lr){5-5}\cmidrule(lr){6-6}\cmidrule(lr){7-7}\cmidrule(lr){8-8}\cmidrule(lr){9-9}\cmidrule(lr){10-10}\cmidrule(lr){11-11} \addlinespace[0.5ex]
              \textbf{DGF}     & \textbf{DGF}             & $0.7088_{0.0003}$      & $0.7981_{0.0004}$      &   $0.7129_{0.0001}$ &    $0.4200_{0.0003}$ &  $0.3751_{0.0008}$  &\bR$0.4191_{0.0038}$ &  $0.4632_{0.0003}$  &   $0.4799_{0.0010}$ &\bR$0.7939_{0.0001}$  \\
              \textbf{GAT}     & \textbf{GAT}             & $0.6535_{0.0000}$      & $0.7724_{0.0000}$      &   $0.7938_{0.0001}$ &   $0.3751_{0.0018}$  & $0.3570_{0.0037}$   & $0.3134_{0,0013}$   &  $0.5441_{0.0005}$  &  $0.4590_{0.0039}$  &  $0.7458_{0.0002}$   \\
              \textbf{DGF+GAT} & \textbf{DGF}             & $0.7182_{0.0003}$      & $0.0.8106_{0.0000}$    & $ 0.8110_{0.0000}$  &$0.38577_{0.0024}$    &  $0.3009_{0.0064}$  & $0.3173_{0.0043}$   &  $0.5523_{0.0001}$  &  $0.5257_{0.0021}$  & $0.7648_{0.0003}$    \\
              \textbf{DGF+GAT} & \textbf{DGF+GAT}         &\bR$0.7322_{0.0002}$    & \bR$0.8200_{0.0004}$   &\bR$0.8202_{0.0000}$ & \bR$0.4594_{0.0000}$ &\bR$0.4225_{0.0004}$ &  $0.3870_{0.0099}$  &\bR$0.5577_{0.0005}$ &\bR$0.5685_{0.0016}$ &  $0.7544_{0.0003}$   \\
              \bottomrule
          \end{tabular}%
      }
  \caption{We look at different GNN designs on a wider set of design spaces. 128 samples are used to train, and tested on the entire NAS space. We refer to DGF+GAT as Ensemble.}
  \label{tab:module_design2} 
  \end{table}

\begin{table}[t!]
  \centering
  \resizebox{\linewidth}{!}{%
\begin{tabular}{lcccHHccc}\toprule
\multicolumn{9}{c}{NASBench-201}\\ \toprule
Timesteps & DGF & Leaky & KQV & Operation & Attention & \multicolumn{3}{c}{Number Of Samples}\\
 & Residual  & ReLU & Projection & Attention  &  Residual & 8 & 16 & 32 \\ \cmidrule(lr){1-1} \cmidrule(lr){2-2} \cmidrule(lr){3-3} \cmidrule(lr){4-4}\cmidrule(lr){5-5} \cmidrule(lr){6-6} \cmidrule(lr){7-7}\cmidrule(lr){8-8}\cmidrule(lr){9-9} \addlinespace[0.5ex]
1 & \ding{51} & \ding{55} & \ding{51} & \ding{51} & \ding{51} & 0.5550 & 0.6265 & 0.6850\\
1 & \ding{51} & \ding{55} & \ding{55} & \ding{51} & \ding{51} & 0.5415 & 0.6074 & 0.6767\\
1 & \ding{55} & \ding{51} & \ding{51} & \ding{51} & \ding{51} & 0.5425 & 0.6127 & 0.6841\\
1 & \ding{51} & \ding{51} & \ding{51} & \ding{51} & \ding{51} & 0.5437 & 0.6115 & 0.6830\\
1 & \ding{55} & \ding{55} & \ding{51} & \ding{51} & \ding{51} & 0.5529 & 0.6200 & 0.6773\\
1 & \ding{51} & \ding{51} & \ding{55} & \ding{51} & \ding{51} & 0.5563 & 0.6100 & 0.6766\\
1 & \ding{55} & \ding{55} & \ding{55} & \ding{51} & \ding{51} & 0.5460 & 0.5883 & 0.6886\\
1 & \ding{55} & \ding{51} & \ding{55} & \ding{51} & \ding{51} & 0.5303 & 0.5864 & 0.6886\\
2 & \ding{51} & \ding{51} & \ding{51} & \ding{51} & \ding{51} & 0.5295 & 0.6173 & 0.6796\\
2 & \ding{51} & \ding{51} & \ding{55} & \ding{51} & \ding{51} & 0.5431 & 0.6025 & 0.6847\\
2 & \ding{51} & \ding{55} & \ding{51} & \ding{51} & \ding{51} & 0.5452 & 0.6284 & 0.6800\\
2 & \ding{51} & \ding{55} & \ding{55} & \ding{51} & \ding{51} & 0.5545 & 0.5993 & 0.6758\\
2 & \ding{55} & \ding{51} & \ding{51} & \ding{51} & \ding{51} & 0.5512 & 0.6204 & 0.6781\\
2 & \ding{55} & \ding{51} & \ding{55} & \ding{51} & \ding{51} & 0.5396 & 0.5906 & 0.6807\\
2 & \ding{55} & \ding{55} & \ding{51} & \ding{51} & \ding{51} & 0.5280 & 0.6207 & 0.6781\\
2 & \ding{55} & \ding{55} & \ding{55} & \ding{51} & \ding{51} & 0.5470 & 0.5945 & \bR$0.6956$\\
3 & \ding{51} & \ding{51} & \ding{51} & \ding{51} & \ding{51} & 0.5488 & 0.6314 & 0.6849\\
3 & \ding{51} & \ding{51} & \ding{55} & \ding{51} & \ding{51} & \bR$0.5644$ & 0.6058 & 0.6751\\
3 & \ding{51} & \ding{55} & \ding{55} & \ding{51} & \ding{51} & 0.5529 & 0.5994 & 0.6806\\
3 & \ding{51} & \ding{55} & \ding{51} & \ding{51} & \ding{51} & 0.5457 & \bR$0.6340$ & 0.6863\\
3 & \ding{55} & \ding{51} & \ding{55} & \ding{51} & \ding{51} & 0.5579 & 0.5999 & 0.6909\\
3 & \ding{55} & \ding{55} & \ding{51} & \ding{51} & \ding{51} & 0.5429 & 0.6321 & 0.6874\\
3 & \ding{55} & \ding{51} & \ding{51} & \ding{51} & \ding{51} & 0.5372 & 0.6279 & 0.6835\\
3 & \ding{55} & \ding{55} & \ding{55} & \ding{51} & \ding{51} & 0.5436 & 0.5955 & 0.6862\\ \bottomrule
\end{tabular}

\begin{tabular}{lcccHHccc}\toprule
\multicolumn{9}{c}{PNAS}\\ \toprule
Timesteps & DGF & Leaky & KQV & Operation & Attention & \multicolumn{3}{c}{Number Of Samples}\\
 & Residual  & ReLU & Projection & Attention  &  Residual & 8 & 16 & 32 \\ \cmidrule(lr){1-1} \cmidrule(lr){2-2} \cmidrule(lr){3-3} \cmidrule(lr){4-4}\cmidrule(lr){5-5} \cmidrule(lr){6-6} \cmidrule(lr){7-7}\cmidrule(lr){8-8}\cmidrule(lr){9-9} \addlinespace[0.5ex]
1 & \ding{51} & \ding{55} & \ding{55} & \ding{51} & \ding{51} & 0.1887 & 0.3354 & 0.4763\\
1 & \ding{55} & \ding{55} & \ding{51} & \ding{51} & \ding{51} & 0.1436 & 0.2878 & 0.3994\\
1 & \ding{51} & \ding{55} & \ding{51} & \ding{51} & \ding{51} & 0.1612 & 0.3129 & 0.4290\\
1 & \ding{55} & \ding{51} & \ding{51} & \ding{51} & \ding{51} & 0.1526 & 0.2919 & 0.4178\\
1 & \ding{55} & \ding{55} & \ding{55} & \ding{51} & \ding{51} & 0.2085 & 0.3242 & 0.4546\\
1 & \ding{51} & \ding{51} & \ding{55} & \ding{51} & \ding{51} & 0.1823 & 0.3417 & \bR$0.4837$\\
1 & \ding{51} & \ding{51} & \ding{51} & \ding{51} & \ding{51} & 0.1472 & 0.3096 & 0.4467\\
1 & \ding{55} & \ding{51} & \ding{55} & \ding{51} & \ding{51} & 0.2031 & 0.3121 & 0.4540\\
2 & \ding{51} & \ding{51} & \ding{55} & \ding{51} & \ding{51} & 0.2417 & 0.3568 & 0.4666\\
2 & \ding{51} & \ding{55} & \ding{55} & \ding{51} & \ding{51} & 0.2448 & 0.3617 & 0.4662\\
2 & \ding{51} & \ding{51} & \ding{51} & \ding{51} & \ding{51} & 0.1804 & 0.3212 & 0.4760\\
2 & \ding{55} & \ding{51} & \ding{55} & \ding{51} & \ding{51} & 0.2337 & 0.3448 & 0.4397\\
2 & \ding{51} & \ding{55} & \ding{51} & \ding{51} & \ding{51} & 0.1822 & 0.3238 & 0.4583\\
2 & \ding{55} & \ding{55} & \ding{55} & \ding{51} & \ding{51} & 0.2183 & 0.3263 & 0.4368\\
2 & \ding{55} & \ding{55} & \ding{51} & \ding{51} & \ding{51} & 0.1912 & 0.3091 & 0.4516\\
2 & \ding{55} & \ding{51} & \ding{51} & \ding{51} & \ding{51} & 0.1866 & 0.3125 & 0.4406\\
3 & \ding{51} & \ding{55} & \ding{55} & \ding{51} & \ding{51} & 0.2523 & \bR$0.3639$ & 0.4541\\
3 & \ding{55} & \ding{55} & \ding{55} & \ding{51} & \ding{51} & 0.2271 & 0.3511 & 0.4496\\
3 & \ding{55} & \ding{55} & \ding{51} & \ding{51} & \ding{51} & 0.1802 & 0.3199 & 0.4477\\
3 & \ding{51} & \ding{51} & \ding{55} & \ding{51} & \ding{51} & \bR$0.2586$ & 0.3488 & 0.4606\\
3 & \ding{55} & \ding{51} & \ding{55} & \ding{51} & \ding{51} & 0.2100 & 0.3438 & 0.4442\\
3 & \ding{55} & \ding{51} & \ding{51} & \ding{51} & \ding{51} & 0.1814 & 0.3107 & 0.4415\\
3 & \ding{51} & \ding{55} & \ding{51} & \ding{51} & \ding{51} & 0.1770 & 0.3316 & 0.4660\\
3 & \ding{51} & \ding{51} & \ding{51} & \ding{51} & \ding{51} & 0.1894 & 0.3309 & 0.4581\\\bottomrule
\end{tabular}%
}
%
\caption{Results of architecture design ablation. Tested on 1000 randomly sampled architectures. Average over 3 trials. Results depict the Kendall-$\tau$ Correlation of \GN{} with different DGF GAT module implementations.}
\label{tab:nb201_arch_design} 
\end{table}

\subsection{Architecture Design Ablation}
\label{subsec:archdesignabl}


In this section, we take a deep look at key architectural decisions and how they impact the sample efficiency of predictors. Table \ref{tab:transposed_combined_nb101_kdt} reproduces prior work \citep{tagates} experimental setting and looks at the impact of 'Timesteps' (TS), 'Residual Connection' (RS), 'Zero Cost Symmetry Breaking' (ZCSB) and 'Architectural Zero Cost Proxy' (AZCP). We find that residual connection 'RS' has a major impact on KDT, causing a dip from 0.66 to 0.59 on the test indicated by 1\% of NASBench-101. We extend this experimental setting to Table \ref{tab:nb201_arch_design}, where we study the impact of time-steps on \textbf{the entire search space}. We find that in 66\% of cases, more than 1 time-step has a positive impact on accuracy. It is important to note that the impact is lesser than residual connection.
\\
\\

\begin{table}[h!]
    \centering
    \resizebox{0.7\linewidth}{!}{%
    \begin{tabular}{cccccccccccccccc}\toprule
    & \multicolumn{8}{c}{1\% of NASBench-101} & \multicolumn{7}{c}{5\% of NASBench-101} \\ \cmidrule(lr){1-1} \cmidrule(lr){2-9} \cmidrule(lr){10-16} \addlinespace[0.5ex]
    TS & 1 & 2 & 3 & 1 & 2 & 2 & 2 & 2 & 1 & 2 & 3 & 1 & 1 & 1 & 1 \\  \cmidrule(lr){1-1} \cmidrule(lr){2-9} \cmidrule(lr){10-16} \addlinespace[0.5ex]
    RS & \cmark & \cmark & \cmark & \cmark & \cmark & \cmark & \cmark & \xmark & \cmark & \cmark & \cmark & \cmark & \cmark & \cmark & \cmark \\
    ZCSB & \cmark & \cmark & \cmark & \cmark & \cmark & \xmark & \xmark & \cmark & \cmark & \cmark & \cmark & \cmark & \xmark & \xmark & \cmark \\
    AZCP & \xmark & \xmark & \xmark & \cmark & \cmark & \cmark & \xmark & \cmark & \xmark & \xmark & \xmark & \cmark & \xmark & \cmark & \cmark \\  \cmidrule(lr){1-1} \cmidrule(lr){2-9} \cmidrule(lr){10-16} \addlinespace[0.5ex]
    KDT & 0.65 & 0.67 & 0.66 & \bR$0.68$ & 0.66 & 0.65 & 0.65 & 0.59 & 0.78 & 0.76 & 0.76 & \bR$0.79$ & 0.76 & 0.78 & 0.78 \\
    \bottomrule
    \end{tabular}%
    }
    \caption{Ablation for training on x\% of 7290 samples on NB101, and testing on 7290 samples on NB101.}
    \label{tab:transposed_combined_nb101_kdt} 
\end{table}

Finally, we conduct a large scale study on NASBench-101, NASBench-201 and PNAS. In this study, we look at the impact of having a 'DGF residual', 'GAT LeakyReLU' and 'GAT KQV-Projection'. From Equation \ref{eq:attention-coefficients}, we can see that the projection matrix $W_{p}$ is shared, whereas in typical attention mechanism, we have different projection matrices for the key, query and value tensors. 

Thus, using the 'KQV-Projection' implies using $W_{qp}$, $W_{kp}$ and $W_{vp}$ matrices as follows:

\begin{equation}
\text{Attn}_{j}(X^{l}) = \text{softmax}(\text{LeakyReLU}(A_{j} \cdot a(W^{l}_{qp}X^{l} \cdot W_{kp}X^{l}_{j}))) \cdot W_{vp}X^{l}_{j}
\label{eq:attention-coefficientskqv_abl}
\end{equation}
\begin{equation}
X^{l+1} = \text{{LayerNorm}}\left( \sigma(OW^{l}_{o}) \odot \sum_{j=1}^{n} \text{Attn}_{j}(X^{l}) \right)
\label{eq:graph-attention-outputkqv_abl}
\end{equation}

\subsection{Predictor Training and Transfer on all NAS Search Spaces}
\label{subsec:all_spaces_predictor_abl}

In Table \ref{tab:training_from_scratch_allspaces}, we provide the result of training \GN{} on all NAS spaces for a range of sample sizes. Note that in this table, we provide results over \textbf{all neural network architectures} in the NAS benchmark space, for a total of 1487731 neural networks. On NASBench-301, we do not provide ZCP and CAZ results, as we do not compute the zero cost proxy for the million NN architectures we rank on that space. 

\subsection{End-to-end NAS on all NAS Search Spaces}
\label{subsec:all_spaces_nas_abl}

\begin{figure}[t!]
    \centering
    \includegraphics[width=0.95\columnwidth]{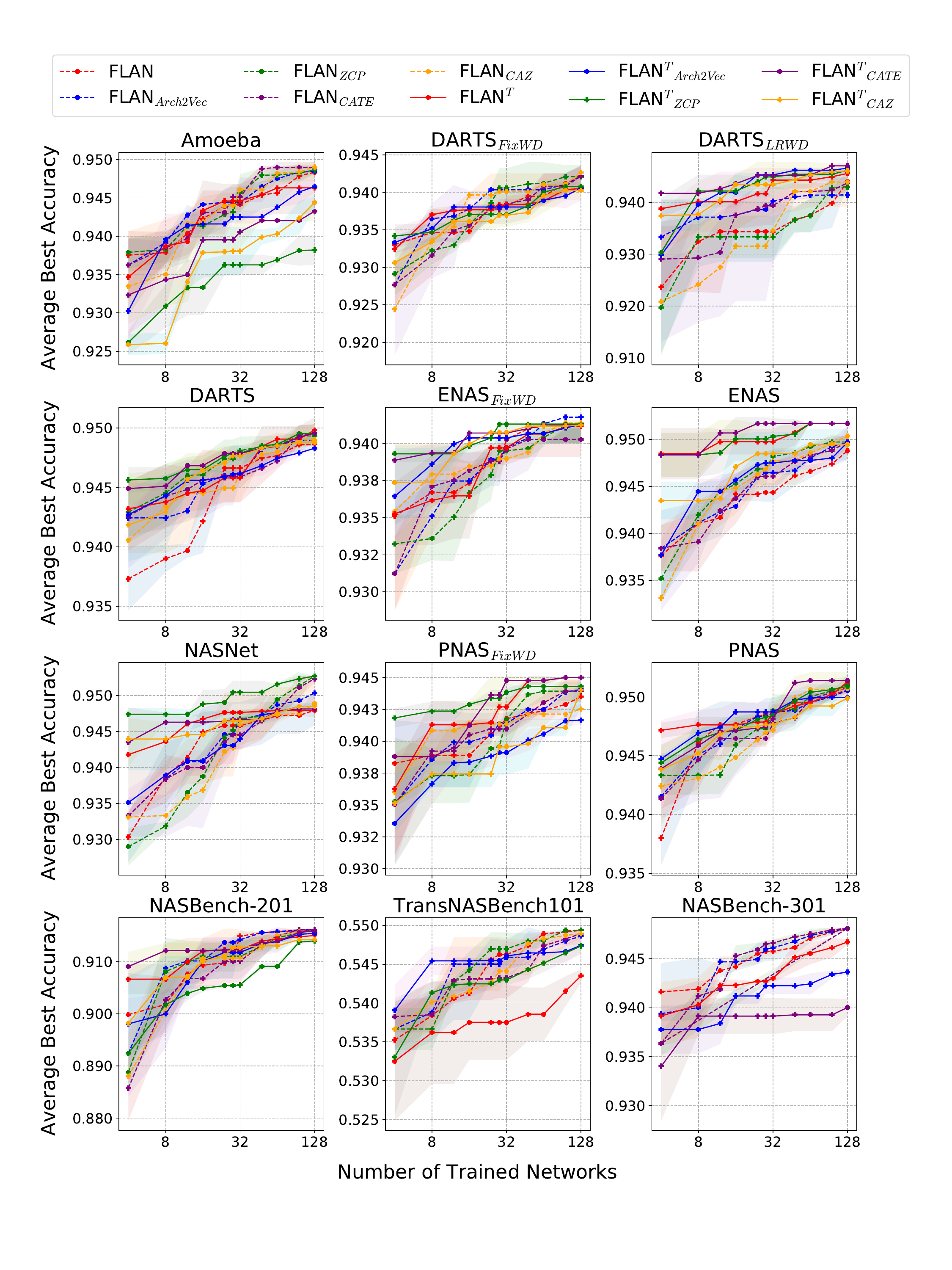}
    \caption{Neural Architecture Search on all NAS spaces detailed in the paper. Accuracies normalized 0-1 except NB201, NB301. Since we evaluate NB301 on 1 million NNs, NB301 does not have ZCP. NB101 in Figure \ref{fig:tagates_sampeff_apndx}. Source-target search space pairs in Table \ref{tab:all_ss_spaces_mapping}. TB101 task is class\_scene. Search is conducted over all available networks in the NAS space.}
    \label{fig:nas_all_spaces}
\end{figure}

In Figure \ref{fig:nas_all_spaces}, we provide the NAS results for a range of samples and subset of representations on all 13 NAS spaces. In these figures, we provide results over \textbf{all neural network architectures} in the NAS benchmark space, for a total of 1487731 neural networks. 

Further, there is a 'Test Acc.' mismatch between Table \ref{tab:nas1_comp} and Figure \ref{fig:nas_results} for NASBench-201. This is because all our NASBench-201 results use TA-GATES style calculation for determining the validation accuracy. For the NASBench-201 test detailed in Table \ref{tab:nas1_comp}, we use GENNAPEs methodology of calculating the test accuracy. To report our results in the setting specified by GENNAPE, we identify the networks discovered, and average their ori-test accuracy.

\begin{figure}[ht!]
    \centering
    \begin{minipage}{.6\textwidth}
        \begin{lstlisting}
    acc_results = sum([nb2_api.get_more_info(arch_index, 'cifar10-valid', None, use_12epochs_result=False, is_random=seed)['valid-accuracy'] for seed in [777, 888, 999]])/3.
        \end{lstlisting}
        \caption{TA-GATES validation accuracy calculation.}
    \end{minipage}\hfill
    \begin{minipage}{.35\textwidth}
        \begin{lstlisting}
    acc_results = nb2_api.query_meta_info_by_index(arch_index).get_metrics('cifar10-valid', 'ori-test')['accuracy']
        \end{lstlisting}
        \caption{GENNAPE test accuracy calculation.}
    \end{minipage}
\end{figure}

\begin{table}[h]
  \centering
  \resizebox{0.85\linewidth}{!}{%
\begin{tabular}{llllllllll}
\toprule
Search Space &     Predictor &              4 &             8 &            16 &            32 &            64 &           128 &           256 &           512 \\
\midrule
      \multirow{5}{*}{NB101} &               FLAN &  $0.11_{0.05}$ & $0.43_{0.00}$ & $0.53_{0.00}$ & $0.55_{0.00}$ & $0.67_{0.00}$ & $0.73_{0.00}$ & $0.77_{0.00}$ & $0.81_{0.00}$ \\
        &      FLAN$_{CAZ}$ &  $0.32_{0.06}$ & $0.31_{0.01}$ & $0.49_{0.00}$ & $0.60_{0.00}$ & $0.66_{0.00}$ & $0.74_{0.00}$ & $0.78_{0.00}$ & $0.81_{0.00}$ \\
        & FLAN$_{Arch2Vec}$ &  $0.34_{0.03}$ & $0.35_{0.06}$ & $0.43_{0.00}$ & $0.52_{0.00}$ & $0.58_{0.00}$ & $0.67_{0.00}$ & $0.76_{0.00}$ & $0.79_{0.00}$ \\
        &     FLAN$_{CATE}$ &  $0.24_{0.05}$ & $0.44_{0.00}$ & $0.48_{0.00}$ & $0.59_{0.00}$ & $0.65_{0.00}$ & $0.73_{0.00}$ & $0.78_{0.00}$ & $0.81_{0.00}$ \\
        &      FLAN$_{ZCP}$ &  $0.30_{0.01}$ & $0.40_{0.02}$ & $0.51_{0.00}$ & $0.64_{0.00}$ & $0.70_{0.00}$ & $0.76_{0.00}$ & $0.79_{0.00}$ & $0.82_{0.00}$ \\
     \midrule \multirow{5}{*}{NB201} &               FLAN &  $0.28_{0.05}$ & $0.42_{0.03}$ & $0.59_{0.00}$ & $0.64_{0.00}$ & $0.77_{0.00}$ & $0.81_{0.00}$ & $0.86_{0.00}$ & $0.89_{0.00}$ \\
        &      FLAN$_{CAZ}$ &  $0.31_{0.05}$ & $0.42_{0.03}$ & $0.54_{0.01}$ & $0.63_{0.00}$ & $0.75_{0.00}$ & $0.82_{0.00}$ & $0.87_{0.00}$ & $0.89_{0.00}$ \\
        & FLAN$_{Arch2Vec}$ &  $0.29_{0.05}$ & $0.45_{0.04}$ & $0.59_{0.00}$ & $0.63_{0.00}$ & $0.76_{0.00}$ & $0.81_{0.00}$ & $0.87_{0.00}$ & $0.89_{0.00}$ \\
        &     FLAN$_{CATE}$ &  $0.26_{0.04}$ & $0.45_{0.03}$ & $0.54_{0.01}$ & $0.63_{0.00}$ & $0.76_{0.00}$ & $0.82_{0.00}$ & $0.86_{0.00}$ & $0.89_{0.00}$ \\
        &      FLAN$_{ZCP}$ &  $0.27_{0.08}$ & $0.43_{0.03}$ & $0.57_{0.00}$ & $0.64_{0.00}$ & $0.75_{0.00}$ & $0.82_{0.00}$ & $0.86_{0.00}$ & $0.89_{0.00}$ \\
     \midrule \multirow{5}{*}{TB101} &               FLAN &  $0.47_{0.01}$ & $0.50_{0.01}$ & $0.54_{0.00}$ & $0.67_{0.00}$ & $0.72_{0.00}$ & $0.75_{0.00}$ & $0.79_{0.00}$ & $0.81_{0.00}$ \\
        &      FLAN$_{CAZ}$ &  $0.47_{0.01}$ & $0.57_{0.00}$ & $0.59_{0.00}$ & $0.66_{0.00}$ & $0.73_{0.00}$ & $0.77_{0.00}$ & $0.79_{0.00}$ & $0.81_{0.00}$ \\
        & FLAN$_{Arch2Vec}$ &  $0.50_{0.00}$ & $0.58_{0.00}$ & $0.60_{0.00}$ & $0.64_{0.00}$ & $0.70_{0.00}$ & $0.75_{0.00}$ & $0.78_{0.00}$ & $0.80_{0.00}$ \\
        &     FLAN$_{CATE}$ &  $0.49_{0.00}$ & $0.50_{0.01}$ & $0.55_{0.01}$ & $0.66_{0.00}$ & $0.73_{0.00}$ & $0.75_{0.00}$ & $0.79_{0.00}$ & $0.80_{0.00}$ \\
        &      FLAN$_{ZCP}$ &  $0.44_{0.02}$ & $0.53_{0.01}$ & $0.53_{0.00}$ & $0.67_{0.00}$ & $0.73_{0.00}$ & $0.77_{0.00}$ & $0.80_{0.00}$ & $0.81_{0.00}$ \\
     \midrule \multirow{3}{*}{NB301} &               FLAN &  $0.24_{0.03}$ & $0.32_{0.04}$ & $0.53_{0.01}$ & $0.66_{0.00}$ & $0.76_{0.00}$ & $0.82_{0.00}$ & $0.85_{0.00}$ & $0.88_{0.00}$ \\
       & FLAN$_{Arch2Vec}$ &  $0.24_{0.03}$ & $0.30_{0.04}$ & $0.51_{0.01}$ & $0.64_{0.00}$ & $0.74_{0.00}$ & $0.81_{0.00}$ & $0.83_{0.00}$ & $0.86_{0.00}$ \\
       &     FLAN$_{CATE}$ &  $0.32_{0.01}$ & $0.34_{0.05}$ & $0.57_{0.00}$ & $0.66_{0.00}$ & $0.77_{0.00}$ & $0.82_{0.00}$ & $0.85_{0.00}$ & $0.87_{0.00}$ \\
    \midrule \multirow{5}{*}{NASNet} &               FLAN &  $0.07_{0.00}$ & $0.09_{0.00}$ & $0.12_{0.00}$ & $0.22_{0.00}$ & $0.30_{0.00}$ & $0.40_{0.00}$ & $0.50_{0.00}$ & $0.57_{0.00}$ \\
       &      FLAN$_{CAZ}$ &  $0.07_{0.00}$ & $0.07_{0.00}$ & $0.24_{0.00}$ & $0.38_{0.00}$ & $0.50_{0.00}$ & $0.54_{0.00}$ & $0.60_{0.00}$ & $0.62_{0.00}$ \\
       & FLAN$_{Arch2Vec}$ &  $0.08_{0.00}$ & $0.06_{0.00}$ & $0.12_{0.00}$ & $0.20_{0.00}$ & $0.37_{0.00}$ & $0.43_{0.00}$ & $0.51_{0.00}$ & $0.54_{0.00}$ \\
       &     FLAN$_{CATE}$ &  $0.08_{0.00}$ & $0.07_{0.00}$ & $0.13_{0.00}$ & $0.20_{0.00}$ & $0.35_{0.01}$ & $0.43_{0.00}$ & $0.52_{0.00}$ & $0.53_{0.00}$ \\
       &      FLAN$_{ZCP}$ &  $0.17_{0.02}$ & $0.17_{0.02}$ & $0.25_{0.00}$ & $0.34_{0.01}$ & $0.51_{0.00}$ & $0.53_{0.00}$ & $0.59_{0.00}$ & $0.62_{0.00}$ \\
      \midrule \multirow{5}{*}{PNAS} &               FLAN &  $0.02_{0.00}$ & $0.04_{0.00}$ & $0.10_{0.01}$ & $0.12_{0.00}$ & $0.26_{0.01}$ & $0.43_{0.00}$ & $0.54_{0.00}$ & $0.61_{0.00}$ \\
         &      FLAN$_{CAZ}$ &  $0.05_{0.00}$ & $0.08_{0.01}$ & $0.18_{0.01}$ & $0.33_{0.00}$ & $0.44_{0.00}$ & $0.52_{0.00}$ & $0.59_{0.00}$ & $0.62_{0.00}$ \\
         & FLAN$_{Arch2Vec}$ &  $0.01_{0.00}$ & $0.04_{0.00}$ & $0.08_{0.00}$ & $0.13_{0.01}$ & $0.31_{0.00}$ & $0.42_{0.00}$ & $0.53_{0.00}$ & $0.60_{0.00}$ \\
         &     FLAN$_{CATE}$ &  $0.00_{0.00}$ & $0.04_{0.00}$ & $0.08_{0.00}$ & $0.13_{0.00}$ & $0.33_{0.00}$ & $0.45_{0.00}$ & $0.53_{0.00}$ & $0.60_{0.00}$ \\
         &      FLAN$_{ZCP}$ &  $0.07_{0.00}$ & $0.10_{0.01}$ & $0.21_{0.01}$ & $0.34_{0.00}$ & $0.46_{0.00}$ & $0.54_{0.00}$ & $0.60_{0.00}$ & $0.63_{0.00}$ \\
     \midrule \multirow{5}{*}{DARTS} &               FLAN &  $0.04_{0.00}$ & $0.04_{0.00}$ & $0.08_{0.00}$ & $0.24_{0.01}$ & $0.29_{0.02}$ & $0.51_{0.00}$ & $0.58_{0.00}$ & $0.68_{0.00}$ \\
        &      FLAN$_{CAZ}$ &  $0.11_{0.01}$ & $0.10_{0.00}$ & $0.23_{0.01}$ & $0.40_{0.01}$ & $0.52_{0.00}$ & $0.59_{0.00}$ & $0.64_{0.00}$ & $0.70_{0.00}$ \\
        & FLAN$_{Arch2Vec}$ &  $0.04_{0.00}$ & $0.03_{0.00}$ & $0.10_{0.00}$ & $0.17_{0.01}$ & $0.35_{0.01}$ & $0.49_{0.00}$ & $0.57_{0.00}$ & $0.64_{0.00}$ \\
        &     FLAN$_{CATE}$ &  $0.06_{0.00}$ & $0.03_{0.00}$ & $0.11_{0.01}$ & $0.17_{0.00}$ & $0.37_{0.01}$ & $0.53_{0.00}$ & $0.58_{0.00}$ & $0.67_{0.00}$ \\
        &      FLAN$_{ZCP}$ &  $0.19_{0.02}$ & $0.15_{0.01}$ & $0.29_{0.02}$ & $0.44_{0.00}$ & $0.52_{0.00}$ & $0.62_{0.00}$ & $0.66_{0.00}$ & $0.70_{0.00}$ \\
    \midrule \multirow{5}{*}{Amoeba} &               FLAN &  $0.00_{0.00}$ & $0.03_{0.00}$ & $0.06_{0.00}$ & $0.23_{0.00}$ & $0.29_{0.00}$ & $0.44_{0.00}$ & $0.50_{0.00}$ & $0.61_{0.00}$ \\
       &      FLAN$_{CAZ}$ &  $0.04_{0.00}$ & $0.10_{0.00}$ & $0.22_{0.01}$ & $0.38_{0.00}$ & $0.45_{0.00}$ & $0.54_{0.00}$ & $0.59_{0.00}$ & $0.62_{0.00}$ \\
       & FLAN$_{Arch2Vec}$ &  $0.01_{0.00}$ & $0.05_{0.00}$ & $0.09_{0.00}$ & $0.24_{0.00}$ & $0.31_{0.00}$ & $0.41_{0.00}$ & $0.47_{0.00}$ & $0.56_{0.00}$ \\
       &     FLAN$_{CATE}$ & $-0.00_{0.00}$ & $0.06_{0.00}$ & $0.08_{0.01}$ & $0.27_{0.00}$ & $0.34_{0.00}$ & $0.42_{0.00}$ & $0.51_{0.00}$ & $0.57_{0.00}$ \\
       &      FLAN$_{ZCP}$ &  $0.05_{0.01}$ & $0.09_{0.01}$ & $0.29_{0.01}$ & $0.41_{0.00}$ & $0.47_{0.00}$ & $0.54_{0.00}$ & $0.59_{0.00}$ & $0.62_{0.00}$ \\
      \midrule \multirow{5}{*}{ENAS} &               FLAN &  $0.02_{0.00}$ & $0.00_{0.00}$ & $0.10_{0.00}$ & $0.35_{0.00}$ & $0.32_{0.00}$ & $0.45_{0.00}$ & $0.53_{0.00}$ & $0.64_{0.00}$ \\
         &      FLAN$_{CAZ}$ &  $0.09_{0.01}$ & $0.11_{0.00}$ & $0.26_{0.01}$ & $0.46_{0.00}$ & $0.50_{0.00}$ & $0.56_{0.00}$ & $0.59_{0.00}$ & $0.65_{0.00}$ \\
         & FLAN$_{Arch2Vec}$ &  $0.03_{0.00}$ & $0.02_{0.00}$ & $0.08_{0.00}$ & $0.24_{0.01}$ & $0.32_{0.00}$ & $0.42_{0.00}$ & $0.51_{0.00}$ & $0.56_{0.00}$ \\
         &     FLAN$_{CATE}$ &  $0.03_{0.00}$ & $0.00_{0.00}$ & $0.09_{0.00}$ & $0.34_{0.00}$ & $0.40_{0.00}$ & $0.46_{0.00}$ & $0.54_{0.00}$ & $0.58_{0.00}$ \\
         &      FLAN$_{ZCP}$ &  $0.13_{0.01}$ & $0.11_{0.00}$ & $0.33_{0.01}$ & $0.52_{0.00}$ & $0.54_{0.00}$ & $0.56_{0.00}$ & $0.61_{0.00}$ & $0.65_{0.00}$ \\\midrule
\multirow{5}{*}{ENAS\_fix-w-d} &               FLAN &  $0.01_{0.01}$ & $0.06_{0.01}$ & $0.28_{0.01}$ & $0.41_{0.01}$ & $0.48_{0.00}$ & $0.53_{0.00}$ & $0.60_{0.00}$ & $0.65_{0.00}$ \\
&      FLAN$_{CAZ}$ &  $0.05_{0.01}$ & $0.19_{0.01}$ & $0.36_{0.01}$ & $0.44_{0.00}$ & $0.50_{0.00}$ & $0.55_{0.00}$ & $0.60_{0.00}$ & $0.65_{0.00}$ \\
& FLAN$_{Arch2Vec}$ &  $0.03_{0.00}$ & $0.07_{0.01}$ & $0.29_{0.01}$ & $0.38_{0.01}$ & $0.44_{0.00}$ & $0.49_{0.00}$ & $0.57_{0.00}$ & $0.62_{0.00}$ \\
&     FLAN$_{CATE}$ & $-0.00_{0.01}$ & $0.04_{0.01}$ & $0.29_{0.01}$ & $0.45_{0.00}$ & $0.50_{0.00}$ & $0.55_{0.00}$ & $0.59_{0.00}$ & $0.64_{0.00}$ \\
&      FLAN$_{ZCP}$ &  $0.10_{0.03}$ & $0.20_{0.02}$ & $0.34_{0.00}$ & $0.47_{0.00}$ & $0.50_{0.00}$ & $0.55_{0.00}$ & $0.59_{0.00}$ & $0.64_{0.00}$ \\\midrule
\multirow{5}{*}{PNAS\_fix-w-d} &               FLAN &  $0.05_{0.01}$ & $0.18_{0.01}$ & $0.26_{0.01}$ & $0.31_{0.00}$ & $0.43_{0.00}$ & $0.57_{0.00}$ & $0.61_{0.00}$ & $0.66_{0.00}$ \\
&      FLAN$_{CAZ}$ &  $0.00_{0.00}$ & $0.17_{0.01}$ & $0.28_{0.01}$ & $0.30_{0.01}$ & $0.43_{0.00}$ & $0.57_{0.00}$ & $0.62_{0.00}$ & $0.66_{0.00}$ \\
& FLAN$_{Arch2Vec}$ & $-0.00_{0.00}$ & $0.17_{0.01}$ & $0.27_{0.01}$ & $0.26_{0.01}$ & $0.41_{0.00}$ & $0.55_{0.00}$ & $0.61_{0.00}$ & $0.66_{0.00}$ \\
&     FLAN$_{CATE}$ &  $0.02_{0.00}$ & $0.18_{0.01}$ & $0.28_{0.01}$ & $0.35_{0.01}$ & $0.44_{0.00}$ & $0.59_{0.00}$ & $0.62_{0.00}$ & $0.67_{0.00}$ \\
&      FLAN$_{ZCP}$ &  $0.05_{0.01}$ & $0.18_{0.01}$ & $0.28_{0.00}$ & $0.31_{0.01}$ & $0.44_{0.00}$ & $0.58_{0.00}$ & $0.62_{0.00}$ & $0.66_{0.00}$ \\\midrule
\multirow{5}{*}{DARTS\_fix-w-d} &               FLAN &  $0.08_{0.00}$ & $0.03_{0.01}$ & $0.21_{0.00}$ & $0.29_{0.01}$ & $0.46_{0.00}$ & $0.56_{0.00}$ & $0.62_{0.00}$ & $0.67_{0.00}$ \\
&      FLAN$_{CAZ}$ &  $0.08_{0.00}$ & $0.07_{0.01}$ & $0.19_{0.00}$ & $0.28_{0.01}$ & $0.43_{0.00}$ & $0.56_{0.00}$ & $0.63_{0.00}$ & $0.67_{0.00}$ \\
& FLAN$_{Arch2Vec}$ &  $0.06_{0.00}$ & $0.06_{0.01}$ & $0.20_{0.00}$ & $0.24_{0.00}$ & $0.42_{0.00}$ & $0.53_{0.00}$ & $0.61_{0.00}$ & $0.66_{0.00}$ \\
&     FLAN$_{CATE}$ &  $0.06_{0.00}$ & $0.07_{0.01}$ & $0.18_{0.00}$ & $0.30_{0.01}$ & $0.44_{0.00}$ & $0.55_{0.00}$ & $0.62_{0.00}$ & $0.67_{0.00}$ \\
&      FLAN$_{ZCP}$ &  $0.04_{0.00}$ & $0.05_{0.01}$ & $0.18_{0.00}$ & $0.30_{0.01}$ & $0.47_{0.00}$ & $0.57_{0.00}$ & $0.63_{0.00}$ & $0.67_{0.00}$ \\
\bottomrule
\end{tabular}
}
  \caption{Training FLAN from scratch.}
  \label{tab:training_from_scratch_allspaces}
\end{table}

\begin{table}[h]
  \centering
  \resizebox{0.9\linewidth}{!}{%
  \begin{tabular}{lllllllllll}
\toprule
 Search Space &          Predictor &              0 &             4 &             8 &            16 &            32 &            64 &           128 &           256 &           512 \\
\midrule
\multirow{5}{*}{NB101} &                 FLAN$^{T}$ &  $0.54_{0.00}$ & $0.48_{0.01}$ & $0.54_{0.00}$ & $0.57_{0.00}$ & $0.64_{0.00}$ & $0.63_{0.00}$ & $0.71_{0.00}$ & $0.76_{0.00}$ & $0.80_{0.00}$ \\
 &      FLAN$^{T}$$_{CAZ}$ &  $0.52_{0.00}$ & $0.51_{0.00}$ & $0.52_{0.00}$ & $0.59_{0.00}$ & $0.60_{0.00}$ & $0.65_{0.00}$ & $0.71_{0.00}$ & $0.77_{0.00}$ & $0.80_{0.00}$ \\
 & FLAN$^{T}$$_{Arch2Vec}$ & $-0.02_{0.00}$ & $0.39_{0.01}$ & $0.47_{0.00}$ & $0.51_{0.00}$ & $0.60_{0.00}$ & $0.68_{0.00}$ & $0.72_{0.00}$ & $0.77_{0.00}$ & $0.79_{0.00}$ \\
 &     FLAN$^{T}$$_{CATE}$ &  $0.52_{0.00}$ & $0.45_{0.00}$ & $0.50_{0.00}$ & $0.54_{0.00}$ & $0.61_{0.00}$ & $0.65_{0.00}$ & $0.69_{0.00}$ & $0.75_{0.00}$ & $0.79_{0.00}$ \\
 &      FLAN$^{T}$$_{ZCP}$ &  $0.59_{0.00}$ & $0.52_{0.00}$ & $0.58_{0.00}$ & $0.60_{0.00}$ & $0.62_{0.00}$ & $0.67_{0.00}$ & $0.71_{0.00}$ & $0.77_{0.00}$ & $0.80_{0.00}$ \\\midrule
\multirow{5}{*}{NB201} &                 FLAN$^{T}$ &  $0.61_{0.00}$ & $0.67_{0.00}$ & $0.54_{0.05}$ & $0.74_{0.00}$ & $0.75_{0.00}$ & $0.80_{0.00}$ & $0.85_{0.00}$ & $0.86_{0.00}$ & $0.89_{0.00}$ \\
 &      FLAN$^{T}$$_{CAZ}$ &  $0.63_{0.00}$ & $0.50_{0.01}$ & $0.60_{0.01}$ & $0.71_{0.00}$ & $0.78_{0.00}$ & $0.82_{0.00}$ & $0.86_{0.00}$ & $0.87_{0.00}$ & $0.89_{0.00}$ \\
 & FLAN$^{T}$$_{Arch2Vec}$ &  $0.64_{0.00}$ & $0.52_{0.02}$ & $0.56_{0.00}$ & $0.69_{0.00}$ & $0.74_{0.00}$ & $0.79_{0.00}$ & $0.83_{0.00}$ & $0.87_{0.00}$ & $0.90_{0.00}$ \\
 &     FLAN$^{T}$$_{CATE}$ &  $0.49_{0.00}$ & $0.45_{0.02}$ & $0.55_{0.00}$ & $0.69_{0.00}$ & $0.73_{0.00}$ & $0.80_{0.00}$ & $0.83_{0.00}$ & $0.87_{0.00}$ & $0.89_{0.00}$ \\
 &      FLAN$^{T}$$_{ZCP}$ &  $0.50_{0.00}$ & $0.50_{0.00}$ & $0.51_{0.02}$ & $0.66_{0.00}$ & $0.74_{0.00}$ & $0.79_{0.00}$ & $0.84_{0.00}$ & $0.85_{0.00}$ & $0.89_{0.00}$ \\
\midrule \multirow{3}{*}{TB101} &                 FLAN$^{T}$ &  $0.14_{0.00}$ & $0.44_{0.01}$ & $0.42_{0.03}$ & $0.65_{0.00}$ & $0.69_{0.00}$ & $0.72_{0.00}$ & $0.78_{0.00}$ & $0.80_{0.00}$ & $0.82_{0.00}$ \\
  &      FLAN$^{T}$$_{CAZ}$ &  $-0.08_{0.00}$ & $0.64_{0.00}$ & $0.63_{0.00}$ & $0.65_{0.00}$ & $0.68_{0.00}$ & $0.72_{0.00}$ & $0.76_{0.00}$ & $0.78_{0.00}$ & $0.81_{0.00}$ \\
  & FLAN$^{T}$$_{Arch2Vec}$ &  $0.16_{0.00}$ & $0.50_{0.02}$ & $0.62_{0.00}$ & $0.55_{0.01}$ & $0.71_{0.00}$ & $0.76_{0.00}$ & $0.78_{0.00}$ & $0.79_{0.00}$ & $0.82_{0.00}$ \\
  &      FLAN$^{T}$$_{ZCP}$ &  $0.02_{0.00}$ & $0.62_{0.00}$ & $0.64_{0.00}$ & $0.68_{0.00}$ & $0.72_{0.00}$ & $0.75_{0.00}$ & $0.78_{0.00}$ & $0.81_{0.00}$ & $0.83_{0.00}$ \\
 \midrule \multirow{3}{*}{NB301} &                 FLAN$^{T}$ &  $0.28_{0.00}$ & $0.22_{0.01}$ & $0.29_{0.00}$ & $0.52_{0.00}$ & $0.66_{0.00}$ & $0.76_{0.00}$ & $0.82_{0.00}$ & $0.84_{0.00}$ & $0.86_{0.00}$ \\
   & FLAN$^{T}$$_{Arch2Vec}$ &  $0.24_{0.00}$ & $0.31_{0.00}$ & $0.33_{0.00}$ & $0.50_{0.01}$ & $0.66_{0.00}$ & $0.72_{0.00}$ & $0.80_{0.00}$ & $0.83_{0.00}$ & $0.85_{0.00}$ \\
   &     FLAN$^{T}$$_{CATE}$ &  $0.25_{0.00}$ & $0.21_{0.02}$ & $0.25_{0.01}$ & $0.47_{0.01}$ & $0.66_{0.00}$ & $0.75_{0.00}$ & $0.80_{0.00}$ & $0.83_{0.00}$ & $0.86_{0.00}$ \\\midrule
\multirow{4}{*}{NASNet} &                 FLAN$^{T}$ &  $0.33_{0.00}$ & $0.23_{0.02}$ & $0.30_{0.00}$ & $0.34_{0.00}$ & $0.35_{0.00}$ & $0.44_{0.00}$ & $0.49_{0.00}$ & $0.56_{0.00}$ & $0.60_{0.00}$ \\
& FLAN$^{T}$$_{Arch2Vec}$ &  $0.31_{0.00}$ & $0.29_{0.00}$ & $0.29_{0.01}$ & $0.32_{0.01}$ & $0.34_{0.00}$ & $0.43_{0.00}$ & $0.47_{0.00}$ & $0.52_{0.00}$ & $0.56_{0.00}$ \\
&     FLAN$^{T}$$_{CATE}$ &  $0.36_{0.00}$ & $0.34_{0.00}$ & $0.38_{0.00}$ & $0.32_{0.00}$ & $0.41_{0.00}$ & $0.42_{0.00}$ & $0.47_{0.00}$ & $0.55_{0.00}$ & $0.60_{0.00}$ \\
&      FLAN$^{T}$$_{ZCP}$ &  $0.45_{0.00}$ & $0.39_{0.00}$ & $0.42_{0.00}$ & $0.46_{0.00}$ & $0.45_{0.00}$ & $0.54_{0.00}$ & $0.54_{0.00}$ & $0.57_{0.00}$ & $0.61_{0.00}$ \\
 \midrule \multirow{5}{*}{PNAS} &                 FLAN$^{T}$ &  $0.41_{0.00}$ & $0.32_{0.00}$ & $0.38_{0.00}$ & $0.43_{0.00}$ & $0.46_{0.00}$ & $0.48_{0.00}$ & $0.54_{0.00}$ & $0.59_{0.00}$ & $0.64_{0.00}$ \\
  &      FLAN$^{T}$$_{CAZ}$ &  $0.38_{0.00}$ & $0.32_{0.02}$ & $0.25_{0.00}$ & $0.24_{0.00}$ & $0.37_{0.01}$ & $0.49_{0.00}$ & $0.52_{0.00}$ & $0.60_{0.00}$ & $0.65_{0.00}$ \\
  & FLAN$^{T}$$_{Arch2Vec}$ &  $0.31_{0.00}$ & $0.28_{0.00}$ & $0.32_{0.01}$ & $0.41_{0.00}$ & $0.41_{0.00}$ & $0.51_{0.00}$ & $0.57_{0.00}$ & $0.59_{0.00}$ & $0.65_{0.00}$ \\
  &     FLAN$^{T}$$_{CATE}$ &  $0.43_{0.00}$ & $0.34_{0.01}$ & $0.41_{0.01}$ & $0.42_{0.00}$ & $0.46_{0.00}$ & $0.48_{0.00}$ & $0.53_{0.00}$ & $0.56_{0.00}$ & $0.64_{0.00}$ \\
  &      FLAN$^{T}$$_{ZCP}$ &  $0.51_{0.00}$ & $0.45_{0.00}$ & $0.51_{0.00}$ & $0.52_{0.00}$ & $0.51_{0.00}$ & $0.53_{0.00}$ & $0.56_{0.00}$ & $0.60_{0.00}$ & $0.66_{0.00}$ \\\midrule
\multirow{5}{*}{DARTS} &                 FLAN$^{T}$ &  $0.55_{0.00}$ & $0.45_{0.01}$ & $0.48_{0.00}$ & $0.57_{0.00}$ & $0.54_{0.00}$ & $0.59_{0.00}$ & $0.60_{0.00}$ & $0.67_{0.00}$ & $0.69_{0.00}$ \\
  &      FLAN$^{T}$$_{CAZ}$ &  $0.64_{0.00}$ & $0.60_{0.00}$ & $0.50_{0.01}$ & $0.62_{0.00}$ & $0.58_{0.00}$ & $0.62_{0.00}$ & $0.65_{0.00}$ & $0.66_{0.00}$ & $0.69_{0.00}$ \\
  & FLAN$^{T}$$_{Arch2Vec}$ &  $0.50_{0.00}$ & $0.44_{0.01}$ & $0.48_{0.02}$ & $0.57_{0.00}$ & $0.58_{0.00}$ & $0.58_{0.00}$ & $0.61_{0.00}$ & $0.64_{0.00}$ & $0.69_{0.00}$ \\
  &     FLAN$^{T}$$_{CATE}$ &  $0.60_{0.00}$ & $0.35_{0.02}$ & $0.48_{0.00}$ & $0.53_{0.00}$ & $0.50_{0.01}$ & $0.57_{0.00}$ & $0.62_{0.00}$ & $0.64_{0.00}$ & $0.68_{0.00}$ \\
  &      FLAN$^{T}$$_{ZCP}$ &  $0.60_{0.00}$ & $0.52_{0.01}$ & $0.58_{0.00}$ & $0.62_{0.00}$ & $0.61_{0.00}$ & $0.63_{0.00}$ & $0.67_{0.00}$ & $0.70_{0.00}$ & $0.73_{0.00}$ \\\midrule
\multirow{5}{*}{Amoeba} &                 FLAN$^{T}$ &  $0.46_{0.00}$ & $0.42_{0.00}$ & $0.42_{0.00}$ & $0.47_{0.00}$ & $0.45_{0.01}$ & $0.51_{0.00}$ & $0.56_{0.00}$ & $0.60_{0.00}$ & $0.64_{0.00}$ \\
&      FLAN$^{T}$$_{CAZ}$ &  $0.52_{0.00}$ & $0.48_{0.00}$ & $0.44_{0.00}$ & $0.48_{0.00}$ & $0.49_{0.00}$ & $0.52_{0.00}$ & $0.54_{0.00}$ & $0.62_{0.00}$ & $0.66_{0.00}$ \\
& FLAN$^{T}$$_{Arch2Vec}$ &  $0.51_{0.00}$ & $0.38_{0.00}$ & $0.39_{0.00}$ & $0.51_{0.00}$ & $0.50_{0.00}$ & $0.55_{0.00}$ & $0.62_{0.00}$ & $0.61_{0.00}$ & $0.67_{0.00}$ \\
&     FLAN$^{T}$$_{CATE}$ &  $0.53_{0.00}$ & $0.40_{0.01}$ & $0.38_{0.00}$ & $0.49_{0.00}$ & $0.55_{0.00}$ & $0.57_{0.00}$ & $0.60_{0.00}$ & $0.61_{0.00}$ & $0.65_{0.00}$ \\
&      FLAN$^{T}$$_{ZCP}$ &  $0.55_{0.00}$ & $0.51_{0.00}$ & $0.50_{0.00}$ & $0.55_{0.00}$ & $0.56_{0.00}$ & $0.57_{0.00}$ & $0.61_{0.00}$ & $0.62_{0.00}$ & $0.66_{0.00}$ \\
 \midrule \multirow{5}{*}{ENAS} &                 FLAN$^{T}$ &  $0.47_{0.00}$ & $0.43_{0.00}$ & $0.43_{0.00}$ & $0.48_{0.00}$ & $0.44_{0.00}$ & $0.50_{0.00}$ & $0.53_{0.00}$ & $0.56_{0.00}$ & $0.62_{0.00}$ \\
  &      FLAN$^{T}$$_{CAZ}$ &  $0.29_{0.00}$ & $0.26_{0.00}$ & $0.33_{0.00}$ & $0.31_{0.01}$ & $0.40_{0.00}$ & $0.48_{0.00}$ & $0.55_{0.00}$ & $0.58_{0.00}$ & $0.63_{0.00}$ \\
  & FLAN$^{T}$$_{Arch2Vec}$ &  $0.41_{0.00}$ & $0.38_{0.01}$ & $0.35_{0.01}$ & $0.45_{0.00}$ & $0.47_{0.00}$ & $0.50_{0.00}$ & $0.53_{0.00}$ & $0.55_{0.00}$ & $0.62_{0.00}$ \\
  &     FLAN$^{T}$$_{CATE}$ &  $0.44_{0.00}$ & $0.41_{0.00}$ & $0.42_{0.00}$ & $0.40_{0.01}$ & $0.46_{0.00}$ & $0.51_{0.00}$ & $0.47_{0.00}$ & $0.56_{0.00}$ & $0.61_{0.00}$ \\
  &      FLAN$^{T}$$_{ZCP}$ &  $0.51_{0.00}$ & $0.48_{0.00}$ & $0.51_{0.00}$ & $0.54_{0.00}$ & $0.53_{0.00}$ & $0.55_{0.00}$ & $0.60_{0.00}$ & $0.60_{0.00}$ & $0.64_{0.00}$ \\\midrule
\multirow{5}{*}{ENAS\_fix-w-d} &                 FLAN$^{T}$ &  $0.27_{0.00}$ & $0.26_{0.00}$ & $0.28_{0.01}$ & $0.36_{0.00}$ & $0.41_{0.00}$ & $0.47_{0.00}$ & $0.50_{0.00}$ & $0.53_{0.00}$ & $0.57_{0.00}$ \\
&      FLAN$^{T}$$_{CAZ}$ &  $0.41_{0.00}$ & $0.37_{0.00}$ & $0.37_{0.00}$ & $0.36_{0.00}$ & $0.42_{0.00}$ & $0.50_{0.00}$ & $0.51_{0.00}$ & $0.55_{0.00}$ & $0.60_{0.00}$ \\
& FLAN$^{T}$$_{Arch2Vec}$ &  $0.33_{0.00}$ & $0.27_{0.01}$ & $0.30_{0.01}$ & $0.32_{0.00}$ & $0.30_{0.00}$ & $0.41_{0.00}$ & $0.46_{0.00}$ & $0.52_{0.00}$ & $0.55_{0.00}$ \\
&     FLAN$^{T}$$_{CATE}$ &  $0.33_{0.00}$ & $0.28_{0.00}$ & $0.32_{0.01}$ & $0.38_{0.00}$ & $0.39_{0.00}$ & $0.43_{0.00}$ & $0.47_{0.00}$ & $0.52_{0.00}$ & $0.59_{0.00}$ \\
&      FLAN$^{T}$$_{ZCP}$ &  $0.38_{0.00}$ & $0.38_{0.00}$ & $0.34_{0.01}$ & $0.44_{0.00}$ & $0.44_{0.00}$ & $0.48_{0.00}$ & $0.55_{0.00}$ & $0.55_{0.00}$ & $0.60_{0.00}$ \\\midrule
\multirow{5}{*}{PNAS\_fix-w-d} &                 FLAN$^{T}$ &  $0.42_{0.00}$ & $0.34_{0.00}$ & $0.40_{0.00}$ & $0.33_{0.00}$ & $0.40_{0.00}$ & $0.43_{0.00}$ & $0.48_{0.00}$ & $0.50_{0.00}$ & $0.53_{0.00}$ \\
&      FLAN$^{T}$$_{CAZ}$ &  $0.37_{0.00}$ & $0.36_{0.01}$ & $0.40_{0.00}$ & $0.44_{0.00}$ & $0.47_{0.00}$ & $0.51_{0.00}$ & $0.53_{0.00}$ & $0.59_{0.00}$ & $0.62_{0.00}$ \\
& FLAN$^{T}$$_{Arch2Vec}$ &  $0.28_{0.00}$ & $0.26_{0.00}$ & $0.34_{0.00}$ & $0.39_{0.00}$ & $0.39_{0.00}$ & $0.44_{0.00}$ & $0.52_{0.00}$ & $0.60_{0.00}$ & $0.64_{0.00}$ \\
&     FLAN$^{T}$$_{CATE}$ &  $0.43_{0.00}$ & $0.36_{0.01}$ & $0.36_{0.00}$ & $0.40_{0.00}$ & $0.38_{0.00}$ & $0.45_{0.00}$ & $0.49_{0.00}$ & $0.53_{0.00}$ & $0.56_{0.00}$ \\
&      FLAN$^{T}$$_{ZCP}$ &  $0.44_{0.00}$ & $0.37_{0.01}$ & $0.39_{0.01}$ & $0.43_{0.00}$ & $0.44_{0.00}$ & $0.47_{0.00}$ & $0.47_{0.00}$ & $0.54_{0.00}$ & $0.57_{0.00}$ \\\midrule
\multirow{4}{*}{DARTS\_fix-w-d} &                 FLAN$^{T}$ &  $0.28_{0.00}$ & $0.13_{0.01}$ & $0.29_{0.00}$ & $0.28_{0.00}$ & $0.28_{0.00}$ & $0.36_{0.00}$ & $0.40_{0.00}$ & $0.50_{0.00}$ & $0.56_{0.00}$ \\
& FLAN$^{T}$$_{Arch2Vec}$ &  $0.14_{0.00}$ & $0.11_{0.01}$ & $0.22_{0.00}$ & $0.28_{0.00}$ & $0.31_{0.00}$ & $0.41_{0.00}$ & $0.45_{0.00}$ & $0.58_{0.00}$ & $0.66_{0.00}$ \\
&     FLAN$^{T}$$_{CATE}$ &  $0.23_{0.00}$ & $0.22_{0.00}$ & $0.22_{0.00}$ & $0.30_{0.00}$ & $0.32_{0.00}$ & $0.40_{0.00}$ & $0.41_{0.00}$ & $0.47_{0.00}$ & $0.56_{0.00}$ \\
&      FLAN$^{T}$$_{ZCP}$ &  $0.29_{0.00}$ & $0.23_{0.01}$ & $0.34_{0.01}$ & $0.29_{0.01}$ & $0.36_{0.01}$ & $0.40_{0.00}$ & $0.46_{0.00}$ & $0.53_{0.00}$ & $0.60_{0.00}$ \\
\bottomrule
\end{tabular}
}
  \caption{Transfer Learning of the FLAN predictor. Source spaces are provided in Table \ref{tab:all_ss_spaces_mapping}}
  \label{tab:transfer_allspaces}
\end{table}

\end{document}